\documentclass[preprint,12pt]{elsarticle}

\usepackage{amssymb}
\usepackage{amsmath}
\usepackage{cite}
\usepackage{amsmath,amssymb,amsfonts}
\usepackage{algorithmic}
\usepackage{graphicx}
\usepackage{textcomp}
\usepackage{xcolor}
\usepackage{algorithmic}
\usepackage{latexsym}
\usepackage{amssymb}
\usepackage{amsmath}
\usepackage{amsthm}
\usepackage{booktabs}
\usepackage{enumitem}
\usepackage{graphicx}
\usepackage{color}
\usepackage{graphicx}
\usepackage[ruled,vlined]{algorithm2e}
\usepackage{multirow}
\usepackage[T1]{fontenc}
\usepackage{mwe}    
\usepackage{subfig}
\usepackage{pdfpages}
\usepackage{graphicx}
\usepackage{multicol}
\usepackage{multirow}
\usepackage{footnote}
\usepackage{comment}
\usepackage{parskip}
\usepackage{amsmath}
\usepackage{amssymb}
\usepackage{algorithmic}
\usepackage{url}
\usepackage{amsmath}
\usepackage{amsfonts}
\journal{Knowledge based system}

\begin{document}

\begin{frontmatter}



\title{Robust Active Learning (RoAL): Countering Dynamic Adversaries in Active Learning with Elastic Weight Consolidation}

\author[tue]{Ricky Maulana Fajri}\ead{r.m.fajri@tue.nl}
\author[tue]{Yulong Pei}\ead{y.pei.1@tue.nl}
\author[tue,surrey]{Lu Yin}\ead{l.yin@surrey.ac.uk}
\author[tue]{Mykola Pechenizkiy}\ead{mykola.pechenizkiy@tue.nl}

\cortext[cor1]{Ricky Maulana Fajri}

\affiliation[tue]{organization={Department of Mathematics and Computer Science},
            addressline={Groene Loper 3},
             city={Eindhoven},
            postcode={5612 AE},
             state={North Brabant},
             country={Netherlands}}
\affiliation[surrey]{organization={School of Computer Science and Electronic Engineering},
             addressline={Address Stag Hill, University Campus, Guildford },
             city={Guildford},
             postcode={GU2 7XH},
             state={Surrey},
             country={United Kingdom}}
\vspace{2mm}
\begin{abstract}
Despite significant advancements in active learning and adversarial attacks, the intersection of these two fields remains underexplored, particularly in developing robust active learning frameworks against dynamic adversarial threats. The challenge of developing robust active learning frameworks under dynamic adversarial attacks is critical, as these attacks can lead to catastrophic forgetting within the active learning cycle. This paper introduces Robust Active Learning (RoAL), a novel approach designed to address this issue by integrating Elastic Weight Consolidation (EWC) into the active learning process. Our contributions are threefold: First, we propose a new dynamic adversarial attack that poses significant threats to active learning frameworks. Second, we introduce a novel method that combines EWC with active learning to mitigate catastrophic forgetting caused by dynamic adversarial attacks. Finally, we conduct extensive experimental evaluations to demonstrate the efficacy of our approach. The results show that RoAL not only effectively counters dynamic adversarial threats but also significantly reduces the impact of catastrophic forgetting, thereby enhancing the robustness and performance of active learning systems in adversarial environments.
\end{abstract}


\begin{highlights}
\item Introducing a novel dynamic adversarial attack in active learning iteration. 
\item Showing a phenomenon of catastrophic forgetting in active learning that happens due to dynamic adversarial attack.
\item Proposing a novel active learning method that combines Elastic Weight Consolidation to solve catastrophic forgetting. 
\item Experiments on 6 public datasets show an improved robust accuracy compared to recent state-of-the-art method.
\end{highlights}


\end{frontmatter}

\section{Introduction}
Active learning has emerged as a transformative approach in machine learning, optimizing the training process by selectively querying the most informative data points \citep{Settles2008AnAO}. This method has proven effective across diverse range of domains, including image recognition \citep{Bhattacharya2019ActiveLW,Limberg2019ActiveLF}, medical analysis \citep{Khanal2023MVAALMV,Smailagic2018MedALAA}, graph machine learning \citep{caramalau2021sequential,fajri2024structural}, and Large Language Models (LLM) \citep{koksal2023meal,xiao2023freeal,zhang2023llmaaa}. Although active learning has enhanced model performance across these various fields, the rise of adversarial attacks presents a critical challenge to the reliability of these machine learning applications. Such attacks involve subtly modified inputs crafted to deceive machine learning models into making erroneous predictions, thus exposing vulnerabilities in many advanced algorithms \citep{Szegedy2013, Goodfellow2014}. The increasing prevalence of these attacks underscores the need for robust defenses in real-world applications, highlighting an area where the adaptive capabilities of active learning must be further refined and tested.
\begin{figure}[ht!]
    \includegraphics[width=0.9\linewidth]{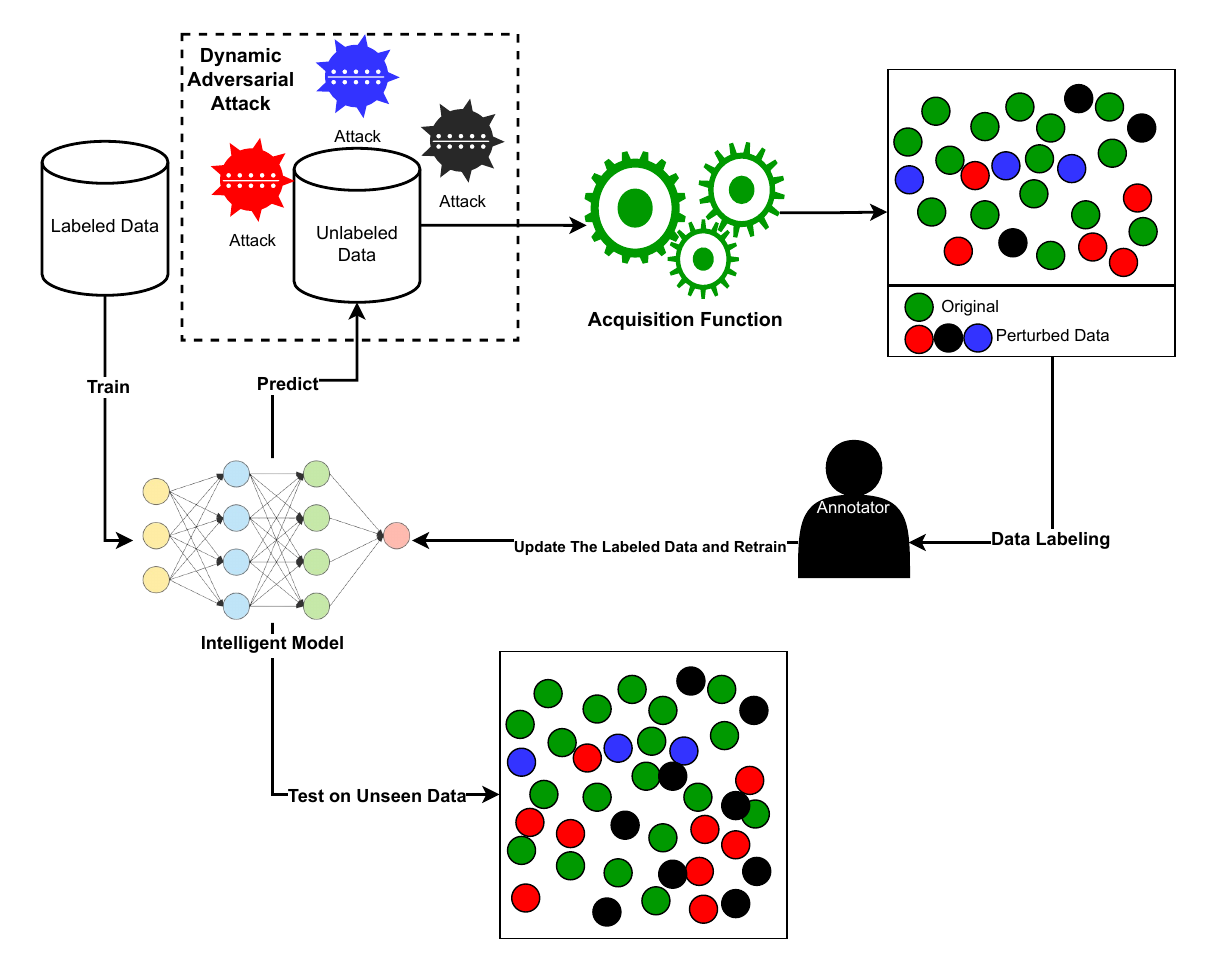}
    \caption{Illustration of Dynamic Adversarial Attack in Active Learning lifecycle}
    \label{fig:illustration}
\end{figure}
As the field of active learning and adversarial attacks continue to evolve, recent studies have begun to connect the field. Various works have attempted to develop active learning frameworks that are robust to adversarial attacks, recognizing the potential for these attacks to undermine active learning processes \citep{DRE, wu2023}. These efforts underscore the critical intersection between improving learning efficiency and ensuring security against malicious inputs.

Despite these advancements, an essential research gap remains in the construction of robust active learning methods under adversarial threats. Most existing approaches assume a static nature of attacks for instance employing Projected Gradient Descent (PGD) uniformly across all iterations of active learning \citep{DRE}. This assumption fails to reflect the dynamic and evolving nature of adversarial attacks in practical scenarios which shown more harmful than traditional adversarial attack \citep{Tao2022DynamicsAwareAA,Chahe2023DynamicAA}. Figure \ref{fig:illustration} illustrates how dynamic adversarial attack occurs in active learning lifecycle. This attack specifically targets the unlabeled data, By perturbing the unlabeled data, the adversary influences the samples chosen for labeling and training, leading to degraded model performance or incorrect predictions as the model updates over time. Moreover, recent work in active learning research under adversarial attacks, such as those documented in \citep{DRE}, have observed catastrophic forgetting , a phenomenon where a model completely loses previously learned information upon acquiring new data. This highlights a substantial need to develop strategies that anticipate and adapt to the evolving tactics of adversaries, thereby dynamically adjusting defensive mechanisms.

To address these challenges, we propose a novel active learning framework that incorporates Elastic Weight Consolidation (EWC) \citep{Kirkpatrick2016OvercomingCF} to mitigate catastrophic forgetting caused by dynamic adversarial conditions. Our approach leverages EWC's ability to prioritize learning important features that are robust against varying attacks. Furthermore, we employ uncertainty sampling as the acquisition function, which has proven to be the most effective method for querying informative data points compared to other acquisition functions. This integration significantly enhances the robustness and efficiency of the active learning process under adversarial conditions.

\textbf{Contributions:} Thus, we summarize the key contributions of the study as follows:
\begin{itemize}
    \item We introduce a dynamic adversarial attack model for active learning, simulating more realistic adversarial behaviors that frequently occur in real-world applications.  
    \item We propose a novel method that optimizes active learning with EWC, specifically tailored to effectively counter catastrophic forgetting in environments characterized by dynamic adversarial threats.
    \item Our extensive experiments demonstrate that active learning, optimized with EWC through uncertainty sampling, notably enhances the model's robustness when facing dynamic adversarial attacks, thereby ensuring more reliable and stable performance across various learning tasks.
\end{itemize}

This paper is organized as follows: Section \ref{section:related_work} reviews related work in the fields of active learning and adversarial attacks. Section \ref{section:problem} describes the problem we are focusing in this study. Section \ref{section:proposed} illustrates the proposed method including the proposed algorithm. Section \ref{section:experiments} and \ref{section:results} present experimental setups and results. Finally, Section \ref{section:conclusion} concludes the paper and suggests directions for future research.
\section{Related work}\label{section:related_work}

Active Learning \citep{Settles2008AnAO} shows interesting results when there is a need to balance the machine learning performance and labeling budget. It has shown many great results in solving the labeling bottleneck in many domains including text classification, image recognition, and large language models. Recently, it has attracted many researchers to combine adversarial sampling attacks for active learning.

The adversarial attacks and active learning research community is divided into two main approaches. The first community believes that it is possible to use the adversarial sample as an acquisition function. Many studies propose a novel approach that uses adversarial attacks as a sampling strategy. For example Mayer and Timofte \citep{Mayer2018AdversarialSF} first introduce the implementation of Adversarial Sampling for Active Learning(ASAL). ASAL is a GAN-based active learning method that generates high entropy samples by searching for similar samples from the pool instead of directly annotating synthetic samples. By synthesizing high entropy samples and retrieving similar real samples from the pool for labeling, ASAL achieves superior performance over random sample selection while maintaining a sub-linear run-time complexity compared to conventional uncertainty sampling methods. Building upon this concept, Generative Adversarial Active Learning (GAAL) \citep{Zhu2017GenerativeAA}, Task-aware Variational Adversarial Active Learning (TA-VAAL) \citep{Sinha2019VariationalAA}, Dual Adversarial network for Active Learning \citep{Wang2020DualAN} delve into Generative Adversarial network for Active Learning. GAAL aims to maximize classification accuracy within a fixed labeling budget, offering a unique strategy to enhance the efficiency and effectiveness of active learning processes, while TA-VAAL combines task-aware active learning with variational adversarial techniques to select informative and influential samples for model training. Finally, DAAL effectively addresses the overlapping problem in uncertainty-based data selection for active learning. 

On the other hand, the second kind of community believes that adversarial attack may harm the active learning performance thus it is essential to develop an active learning framework that is robust under adversarial attack. For instance, Wu et al \citep{wu2023} propose a methodology that not only enhances robustness to adversarial perturbations but also ensures fairness among different demographic groups by employing joint inconsistency metrics. This approach strategically selects data points that optimize performance in both standard and adversarial contexts, emphasizing the importance of fairness in model training. More recently, Guo et al \citep{DRE} introduce an innovative selection method based on the density and entropy of data. This method aims to improve the adversarial robustness of active learning models by prioritizing data points from dense regions with high entropy, which are posited to offer more resilience against crafted adversarial inputs. Our study is aligned with the second community, we believe that it is important to develop a new active learning approach that is robust to adversarial attacks especially when the attack occurs dynamically. 

\section{Problem Formulation}\label{section:problem}
\subsection{Active Learning}
Active learning is an approach where a learning algorithm iteratively selects a subset of unlabeled data, \(\mathcal{U}\), to be labeled by an oracle and added to the labeled dataset, \(\mathcal{L}\), to improve the learning model efficiently. The core of active learning lies in its query function, which can be mathematically represented as follows:
\begin{equation}
Q(\mathcal{U}, f_\theta) = \arg\max_{x \in \mathcal{U}} \Phi(x, f_\theta)
\end{equation}
where \(f_\theta\) denotes the current state of the model parameterized by \(\theta\), and \(\Phi\) represents the querying criterion (e.g., uncertainty). Usually, the most uncertain sample is selected as a candidate for labeling. After selection, the labeled data are integrated into \(\mathcal{L}\) as follows:
\begin{equation}
\mathcal{L} \leftarrow \mathcal{L} \cup \{(x, y) : x \in Q(\mathcal{U}, f_\theta), y = \text{oracle}(x)\}
\end{equation}

\subsection{Adversarial Attacks}
In the context of machine learning, adversarial attacks involve introducing small perturbations, \(\delta\), to input data \(\mathbf{x}\) to produce adversarial examples \(\mathbf{x}_{adv}\), that mislead the model. The perturbation is typically bounded by a norm constraint to ensure subtlety:
\begin{equation}
\mathbf{x}_{adv} = \mathbf{x} + \delta, \quad \text{such that } \|\delta\|_p \leq \epsilon
\end{equation}
For dynamic adversarial attacks within an active learning context, the adversary may modify the data in \(\mathcal{U}\) dynamically in response to the learning model's updates. This can be modeled as:
\begin{equation}
\delta = \arg\max_{\|\delta\|_p \leq \epsilon} L (f_\theta; \mathbf{x} + \delta, y)
\end{equation}
where \( L \) is the loss function used by the model, and the adversarial examples are generated to maximize the model's prediction error under the constraint that the perturbation is small.

In this context, it is important to note that the class label \( y \) from the unlabeled set \( \mathcal{U} \) is typically unknown. Therefore, in adversarial attacks, \( y \) is often estimated or assumed based on the current state of the model \( f_\theta \). This assumption is crucial for generating adversarial examples effectively, as the goal is to create perturbations that lead the model to make incorrect predictions.

\subsection{Objective Function}
To construct a robust active learning system under dynamic adversarial attacks, the objective function should ideally minimize the expected loss over both the adversarial and clean distributions within the labeled and unlabeled datasets. We propose the following formulation:
\begin{align}
& \min_{\theta} \left\{ \mathbb{E}_{(\mathbf{x}, y) \in \mathcal{L}}[L(f_\theta; \mathbf{x}, y)] \right\} \nonumber \\
& + \gamma \mathbb{E}_{\mathbf{x} \in \mathcal{U}} \left[ \max_{\|\delta\|_p \leq \epsilon} L (f_\theta; \mathbf{x} + \delta, \hat{y}(\mathbf{x}+\delta)) \right]
\end{align}
In this function:
\begin{itemize}
  \item The first term \(\mathbb{E}_{(\mathbf{x}, y) \in \mathcal{L}}[L(f_\theta; \mathbf{x}, y)]\) represents the average loss over the clean, labeled data.
  \item The second term captures the expected maximum loss due to adversarial perturbations in the unlabeled set, with \(\gamma\) as a regularization parameter controlling the trade-off between accuracy on clean data and robustness to adversarial manipulation.
  \item \(\hat{y}(\mathbf{x}+\delta)\) denotes the predicted label for the adversarial example, which underscores the anticipation of adaptive adversarial behavior within the active learning cycle.
\end{itemize}

\section{Proposed Method}\label{section:proposed}
The vulnerability of active learning to adversarial attacks can lead to catastrophic forgetting, as targeted perturbations may cause a model to focus solely on new data, neglecting prior knowledge. Implementing EWC in active learning helps mitigate this risk by constraining parameter updates, reinforcing stability, and preserving learned information. This, in turn, supports robust uncertainty sampling, allowing for a more reliable selection of data candidates in the attempt to be more robust under adversarial attack.
\subsection{Elastic Weight Consolidation (EWC)}
Elastic Weight Consolidation (EWC) is a regularization technique designed to prevent catastrophic forgetting by discouraging major changes to critical model parameters. This is achieved by imposing a penalty based on the Fisher Information Matrix (FIM), which represents the importance of each parameter for previously learned tasks. EWC balances stability and plasticity, helping the model retain past knowledge while adapting to new tasks.
The EWC regularization term is defined as follows:
\begin{equation}
R(\theta, \theta^{*}) = \frac{\lambda}{2} \sum_{i=1}^{N} F_i \cdot (\theta_i - \theta^{*}_i)^2
\end{equation}
where:
\begin{itemize}
    \item $\lambda$ is the regularization parameter controlling the penalty's strength.
    \item $F_i$ is the diagonal of the Fisher Information Matrix, measuring the sensitivity of the loss function to changes in the $i^{th}$ parameter.
    \item $\theta^{*}_i$ and $\theta_i$ represent the reference and current parameter values, respectively.
\end{itemize}
The Fisher Information Matrix $F_i$ is calculated by taking the expected squared gradients with respect to the loss function:
\begin{equation}
    F_i = \mathbb{E}\left[ \left( \frac{\partial L_0}{\partial \theta_i} \right)^2 \right]
\end{equation}
where $L_0$ represents the loss function for the initial training task. This Fisher Information guides the penalty in EWC, ensuring significant changes to critical parameters are penalized.

\subsection{Acquisition Function with Uncertainty Sampling}
The acquisition function in active learning helps select the most informative data points from the unlabeled dataset $\mathcal{U}$. This function integrates uncertainty sampling to identify samples with high information value, driving the active learning process. The acquisition function is defined as:
\begin{equation}
A(x) = \Phi(x, f_{\theta})    
\end{equation}
where $\Phi(x, f_{\theta})$ represents the uncertainty of the model about sample $x$, typically quantified by the entropy of the predictive distribution:
\begin{equation}
    \Phi(x, f_{\theta}) = -\sum_c p(c|x, \theta) \log p(c|x, \theta)
\end{equation}
This acquisition function prioritizes the most uncertain samples in $\mathcal{U}$, guiding the active learning process by focusing on data points that can provide the most learning value. This approach supports the robust active learning algorithm by selecting the most uncertain samples, obtaining their labels, and adding them to the labeled dataset $\mathcal{L}$ for further training. 

\subsection{Formal Integration of EWC to Active Learning}
The integration of Elastic Weight Consolidation (EWC) with an active learning framework is designed to enhance neural network training by simultaneously maximizing learning from new data in $\mathcal{U}$ while retaining critical information from previously learned tasks in $\mathcal{L}$. This integration formalizes an objective function that combines the EWC regularization term with the mechanism for updating learning parameters based on new data, without explicitly incorporating the acquisition function into this formulation.

The integrated objective function can be represented as:
\begin{align}
\mathcal{L}(\theta) &= L(\theta; \mathcal{L}, \theta^{*}) + \frac{\lambda}{2} \sum_{i=1}^{N} F_i \cdot (\theta_i - \theta^{*}_i)^2, \\
L(\theta; \mathcal{L}, \theta^{*}) &= \sum_{x \in \mathcal{L}} L(x, \theta) + L_0(\theta),
\end{align}
where $L(\theta; \mathcal{L}, \theta^{*})$ is the loss function that includes new data points added from $\mathcal{U}$ to $\mathcal{L}$, selected based on external criteria related to their information content. $L_0(\theta)$ represents the continuation of learning from previous tasks, critical to the EWC framework.

The EWC regularization still relies on the updated Fisher Information Matrix $F_i$, recalculated to reflect both old and new learning influences:
\begin{equation}
F_i^{new} = \mathbb{E}\left[ \left( \frac{\partial L(\theta; \mathcal{L}, \theta^{*})}{\partial \theta_i} \right)^2 \right],
\end{equation}
The equation above calculates the expected squared gradients based solely on the current labeled data batch $\mathcal{L}$ and the continuing influence of previous tasks, allowing the network to adapt dynamically to new information without the need for manual tuning of decay factors. This methodology ensures that the EWC framework remains effective in managing the trade-off between learning new information and preserving existing knowledge in the active learning cycle. The detailed algorithm of the RoAL method is illustrated in algorithm \ref{alg:proposed_alg}.
\begin{algorithm}
\caption{Robust Active Learning (RoAL) with EWC}
\begin{algorithmic}[1]
\REQUIRE Initial labeled dataset $\mathcal{L}$, initial unlabeled dataset $\mathcal{U}$, number of iterations $T$, number of candidates $C$, EWC regularization parameter $\lambda$, sequence of attacks $\mathcal{A}$
\STATE Initialize model $\theta$ and Fisher information matrix $\mathcal{F}$
\STATE Initialize previous parameters $\theta_{\text{prev}}$
\FOR{each iteration $t$ from $1$ to $T$}
    \STATE Train model $\theta$ on $\mathcal{L}$ with EWC regularization using $\mathcal{F}$ and $\theta_{\text{prev}}$
    \STATE Perform uncertainty sampling on $\mathcal{U}$ to select $C$ candidates.
    \STATE Label the selected candidate samples and add them to $\mathcal{L}$
    \STATE Update Fisher information matrix $\mathcal{F}$ and previous parameters $\theta_{\text{prev}}$
    \STATE Select adversarial attack $\mathcal{A}_t$ for iteration $t$
    \STATE Evaluate model under adversarial attack $\mathcal{A}_t$
    \STATE Remove the labeled candidates from $\mathcal{U}$
\ENDFOR
\end{algorithmic}\label{alg:proposed_alg}
\end{algorithm}

\section{Experiments}\label{section:experiments}
\subsection{Experimental Procedure}
We begin by splitting the dataset into training and test sets, further dividing the training data into labeled and unlabeled subsets. Next, an initial model is trained on the labeled subset of the training data. For each iteration of the active learning process, candidate samples are selected from the unlabeled subset based on their predicted uncertainty using the trained model. Subsequently, a dynamic sequence of adversarial attacks, Such as PGD \citep{PGD}, Jitter \citep{jitter}, Fast Adaptive Boundary (FAB) \citep{FAB}, VNI-FGSM \citep{VNI} (we shorten as VNI), and PGDL2 is applied to a portion of the labeled dataset. PGD iteratively refines adversarial examples by taking steps in the gradient direction, while Jitter adds random noise to input images. FAB targets the decision boundary for minimal perturbations, VNI-FGSM normalizes gradients across iterations, and PGDL2 uses L2 norm-based perturbations for smoother adversarial images.

These attacks are designed to simulate a dynamic adversarial strategy and perturb the labeled samples. The model parameters are then updated using the labeled samples along with their adversarial perturbed counterparts. This iteration process continues until the maximum labeling budget is reached. Following the completion of the active learning iterations, the trained model is evaluated on the unseen test set to assess its accuracy and robustness against adversarial attacks. Additionally, to ensure robustness and obtain statistically meaningful results, the entire experiment is repeated 10 times. Finally, the results from each repetition are aggregated and analyzed to determine the robustness of the active learning framework under dynamic adversarial attacks. 
\subsection{Experiment Settings}
For equal comparison of each experiment, we use same deep neural network architecture. The deep neural network consists of four convolutional layers followed by two fully connected layers. The initial convolutional layers, each with a 3x3 kernel size, extract features from the input images, progressively increasing the depth of extracted features from 3 to 64 channels. After each convolutional layer, Rectified Linear Unit (ReLU) activation functions are applied to introduce non-linearity. Subsequent max-pooling layers with a 2x2 kernel and stride of 2 reduce the spatial dimensions of the feature maps by half, aiding in translation invariance and reducing computational complexity. Following the convolutional layers, the feature maps are flattened and fed into a fully connected layer, which further processes the extracted features. All computations were performed on a machine equipped with a Xeon Platinum 8352V CPU and 24 GB of GPU memory, using PyTorch 2.2.2 for coding and implementation.

For an equal active learning labeling procedure, we use 10 iterations and choose 200 candidate samples from each iteration. Initially, we use 400 samples and labels from the labeled dataset as the initial training set and train the neural network model on these 400 data. Moreover, the RoAL method has only one hyperparameter which is $\lambda$ that behaves as the EWC regularizer. We use $\lambda$ = 0.5 for balancing between the network forgetting and learning at the current task.

For the evaluation procedure, we follow DRE method \citep{DRE} which trains the deep neural network on the entire labeled training set. Then a given acquisition function is implemented to select the candidate for labeling from the unlabeled pool. Subsequently, we generate adversarial examples from the selected data and retrain the DNN using both labeled training data and the generated adversarial examples. Finally, we measure the test accuracy and adversarial robustness of the retrained DNN on the test dataset. 
\subsection{Dataset}
\begin{table*}[]
\centering
\caption{Summary of Datasets.}
\label{tab:datasets}
\begin{tabular}{c|c|c|c}
\hline \hline
\textbf{Dataset} & \textbf{Classes} & \textbf{Image Size} & \textbf{Number of Images} \\ \hline \hline
MNIST & 10 & $28 \times 28$ & 60,000 (train), 10,000 (test) \\ \hline
Fashion MNIST & 10 & $28 \times 28$ & 60,000 (train), 10,000 (test) \\ \hline
CIFAR10 & 10 & $32 \times 32$ & 50,000 (train), 10,000 (test) \\ \hline
CIFAR100 & 100 & $32 \times 32$ & 50,000 (train), 10,000 (test) \\ \hline
SVHN & 10 & Variable & 73,257 (train), 26,032 (test) \\ \hline
Caltech101 & 101 & Variable & 9,144 \\ \hline \hline
\end{tabular}
\end{table*}

We evaluate our proposed approach using six publicly available image classification datasets: \textbf{MNIST} \citep{mnist}, \textbf{Fashion MNIST} \citep{fashion}, \textbf{CIFAR10} and \textbf{CIFAR100} \citep{Krizhevsky2009LearningML}, \textbf{SVHN} \citep{svhn}, and \textbf{Caltech101}\citep{caltech101}. The \textbf{MNIST} dataset consists of 28x28 grayscale images of handwritten digits from 0 to 9. \textbf{Fashion MNIST}, designed as a more challenging replacement for MNIST, includes images of fashion items. \textbf{CIFAR10} features 60,000 32x32 color images across 10 classes, and \textbf{CIFAR100} extends this to 100 classes. \textbf{SVHN} contains images of house numbers taken from Google Street View. \textbf{Caltech101} offers a variety of images across 101 object categories. 
\subsection{Evaluation Metrics}
We measure the accuracy and adversarial robustness(we use robust accuracy interchangeably), of each active learning method. Accuracy is determined by the percentage of correctly classified data across the entire test set, while robust accuracy is calculated as the percentage of correctly classified adversarial examples over the entire adversarial test set. The adversarial test set comprises examples crafted from the clean test set using a predefined adversarial attack. 
Formally the accuracy and robust accuracy can be defined as:
\begin{equation}
\text{Accuracy} = \frac{TP + TN}{TP + TN + FP + FN}    
\end{equation}
where:
\begin{itemize}
    \item \(TP\) is the number of true positives (correctly predicted positive instances),
    \item \(TN\) is the number of true negatives (correctly predicted negative instances),
    \item \(FP\) is the number of false positives (incorrectly predicted positive instances), and
    \item \(FN\) is the number of false negatives (incorrectly predicted negative instances).
\end{itemize}
\begin{equation}
\text{Robust Accuracy} = \frac{TP_{\text{adv}} + TN_{\text{adv}}}{TP_{\text{adv}} + TN_{\text{adv}} + FP_{\text{adv}} + FN_{\text{adv}}}    
\end{equation}
where:
\begin{itemize}
    \item \(TP_{\text{adv}}\) is the number of true positives on adversarial examples,
    \item \(TN_{\text{adv}}\) is the number of true negatives on adversarial examples,
    \item \(FP_{\text{adv}}\) is the number of false positives on adversarial examples, and
    \item \(FN_{\text{adv}}\) is the number of false negatives on adversarial examples.
\end{itemize}
\subsection{Baselines methods}
We compare the RoAL method with various active learning methods, divided into two types. 1. A general active learning method that is not designed for adversarial attacks such as Random, Entropy, EntropyDrop, Bald, and MCP. 2. An active learning method that is specifically designed for adversarial attacks such as DRE. The query strategy of each method is defined as follows:
\begin{itemize}
    \item Random: a simplest and model-free method that each data has an equal probability of being selected.
    \item Entropy \citep{Settles2008AnAO}: Select the sample with the highest uncertainty based on the prediction model.
    \item EntropyDrop \citep{entropyDrop}: Calculates the uncertainty over multiple Bayesian neural networks inferred by the Monte Carlo dropout.
    \item BALD \citep{BALD}: The Bayesian Active Learning by Disagreement(BALD) introduces Gaussian Process Classifiers to select informative data points for model training. By maximizing an objective function that balances predictive uncertainty and model complexity.
    \item MCP \citep{MCP}: The Multiple-Boundary Clustering and Prioritization(MCP) works by clustering test samples into boundary areas based on predicted classes and prioritizing sample selection evenly across these boundaries. By ensuring sufficient samples from each boundary area for reconstruction, MCP enhances the retraining process of deep learning models with limited labeled data.
    \item DRE \citep{DRE}: The Density Robust Entropy(DRE) combines adversarial training with active learning by crafting adversarial examples for labeled data using the PGD attack and training the deep neural network with these perturbed examples.
\end{itemize}
\subsection{Adversarial Attack}
In this study, we strategically employ adversarial attacks both during the training and evaluation phases to comprehensively assess the robustness of the active learning framework. We define dynamic adversarial attacks as a set of diverse adversarial techniques applied at different active learning iterations. During the training phase, we apply adversarial attacks, including PGD, Jitter, FAB, VNI, and PGDL2, to the unlabeled datasets. The order of these attacks is randomized to ensure a dynamic nature of adversarial attacks. However, for reproducibility, we fixed the order as PGD, Jitter, FAB, VNI, PGDL2, PGD, Jitter, FAB, VNI, PGDL2, allowing others to replicate our experiments by following the same sequence.

Moreover, during the evaluation phase, the model's performance is evaluated on the test dataset under the influence of various adversarial attacks. This evaluation scenario closely mimics real-world conditions, where adversaries may attempt to exploit vulnerabilities in the deployed model by crafting adversarial examples. By subjecting the model to adversarial attacks during evaluation, we can measure its robustness and generalization capabilities in the presence of unseen adversarial inputs.
\section{Results}\label{section:results}
\subsection{Accuracy Performance}
\begin{figure*}[htb!]
\centering
\subfloat[MNIST]{\includegraphics[width=.38\textwidth]{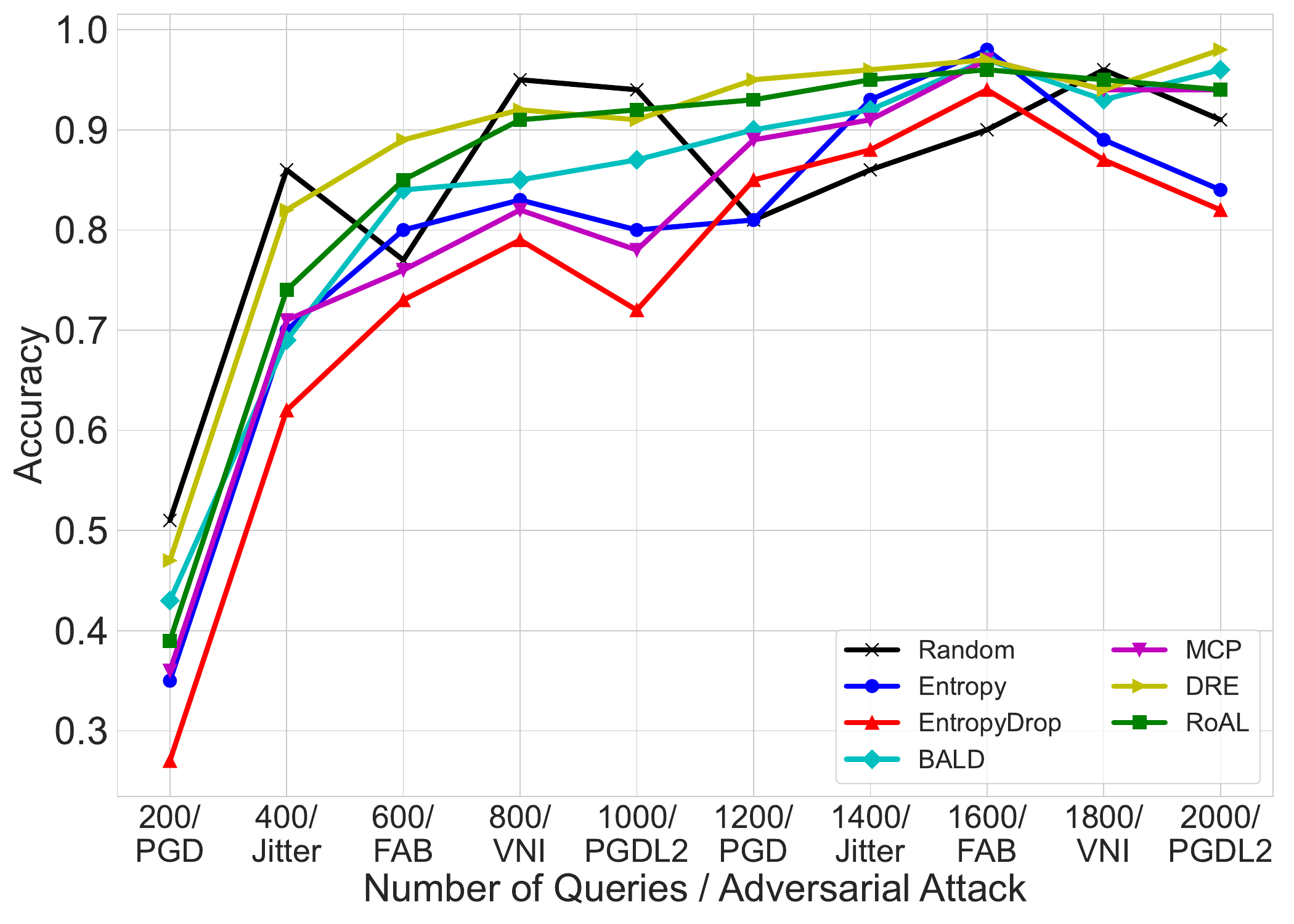}} \hspace{0.01\textwidth}
\subfloat[Fashion MNIST]{\includegraphics[width = .38\textwidth]{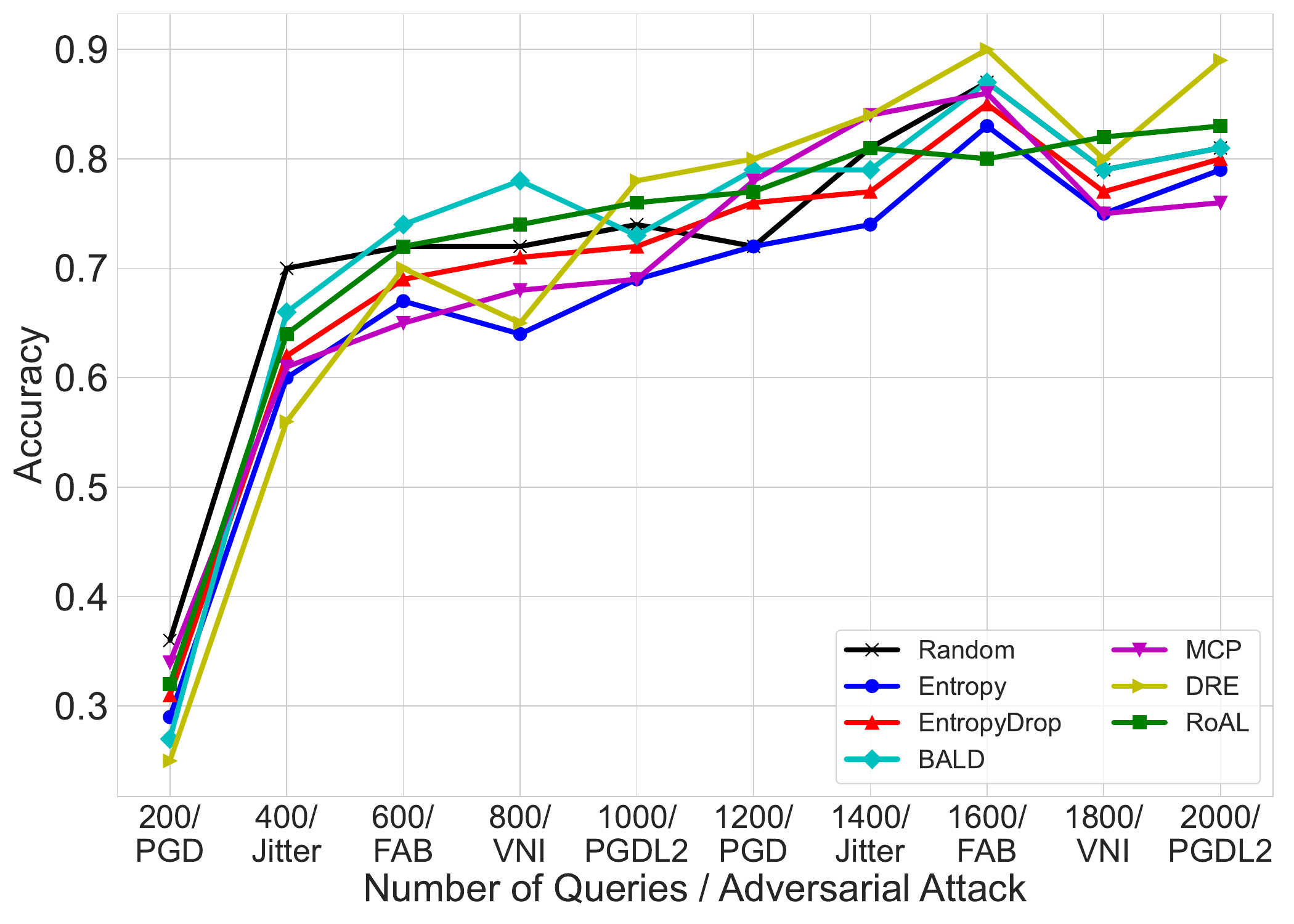}}\hspace{0.01\textwidth} \\
\subfloat[SVHN]{\includegraphics[width =.38\textwidth]{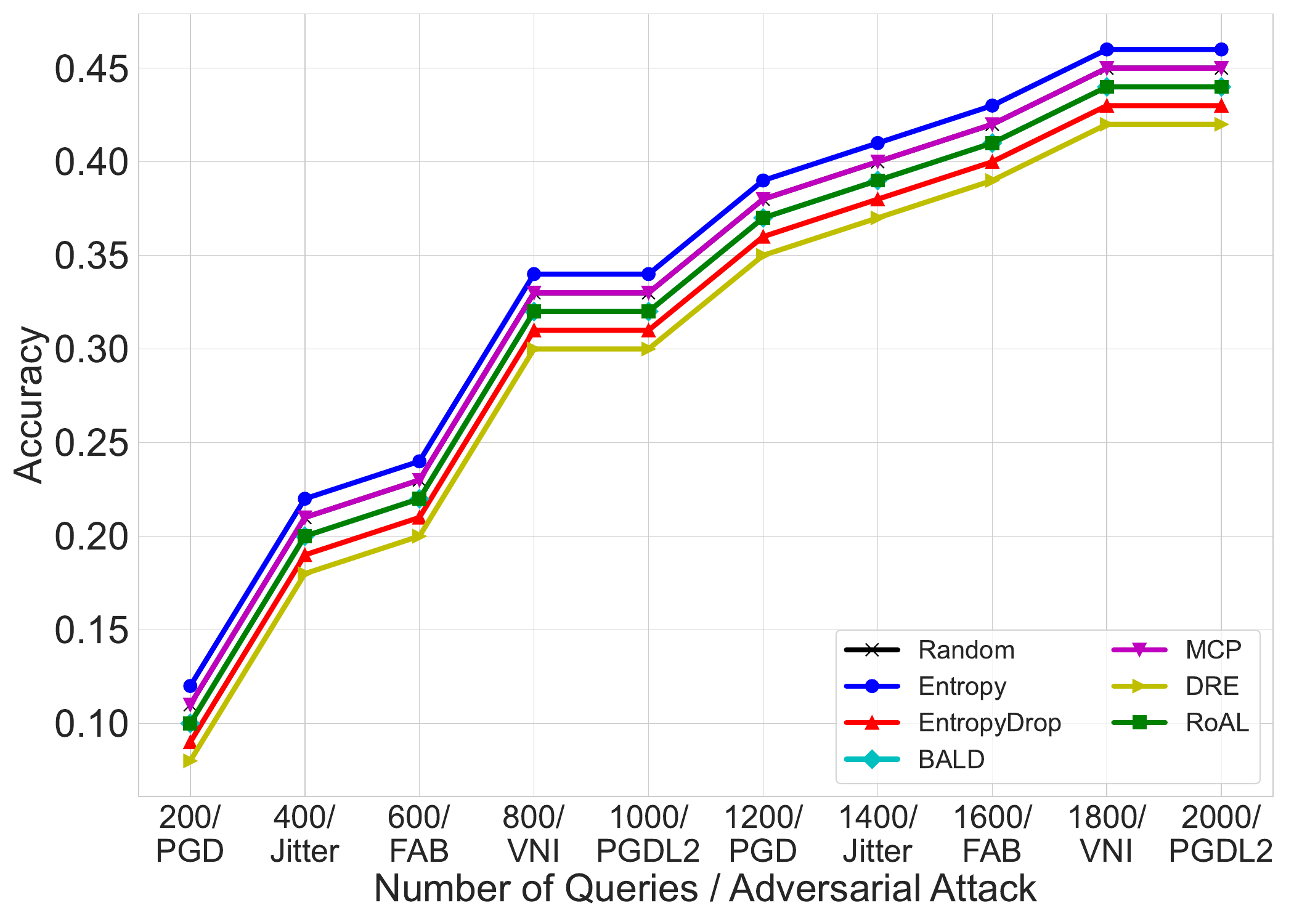}} 
\subfloat[CIFAR10]{\includegraphics[width = .38\textwidth]{cifar10_accuracy.pdf}} \hspace{0.01\textwidth}\\
\subfloat[CIFAR100]{\includegraphics[width = .38\textwidth]{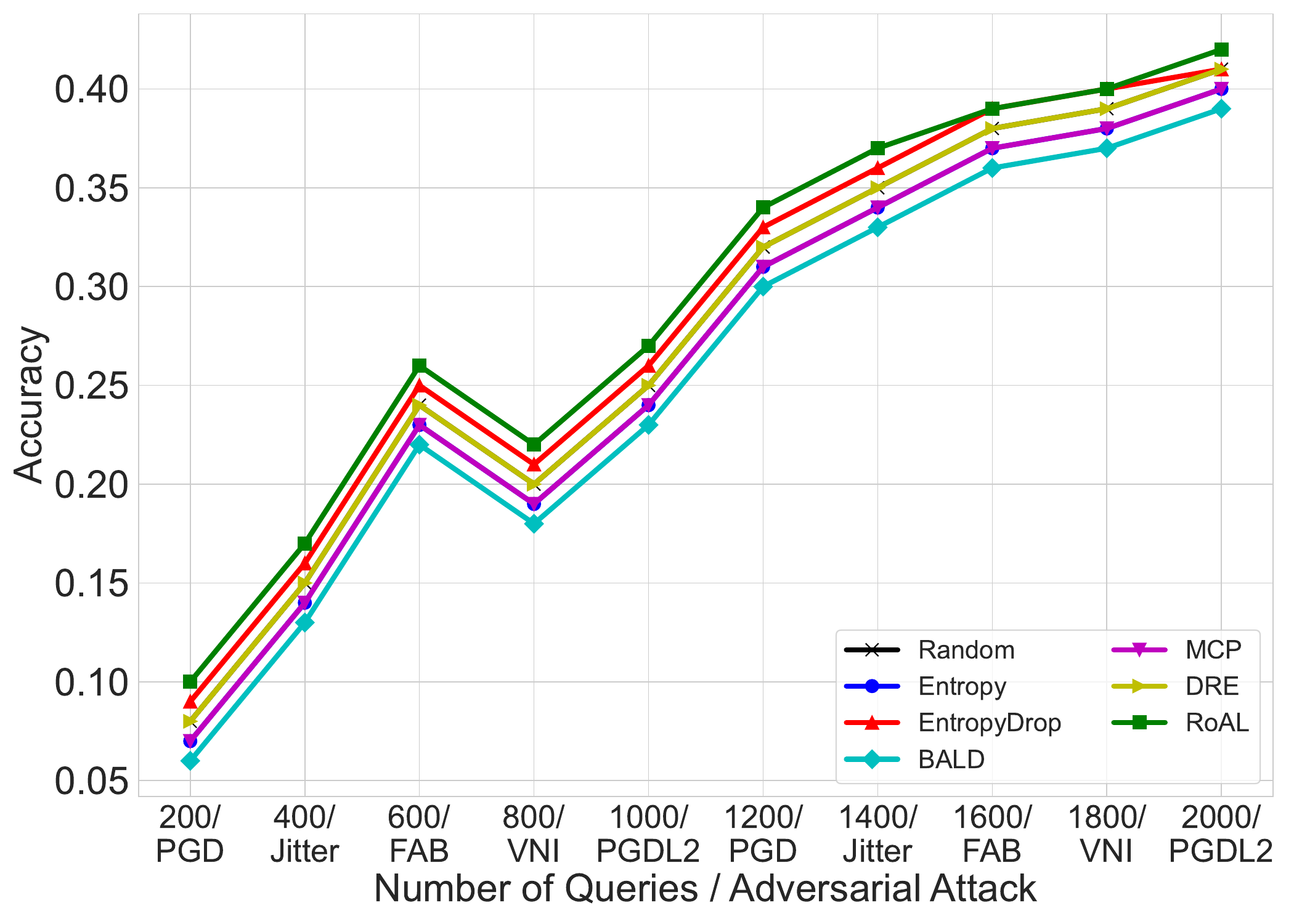}}\hspace{0.01\textwidth}
\subfloat[Caltech101]{\includegraphics[width = .38\textwidth]{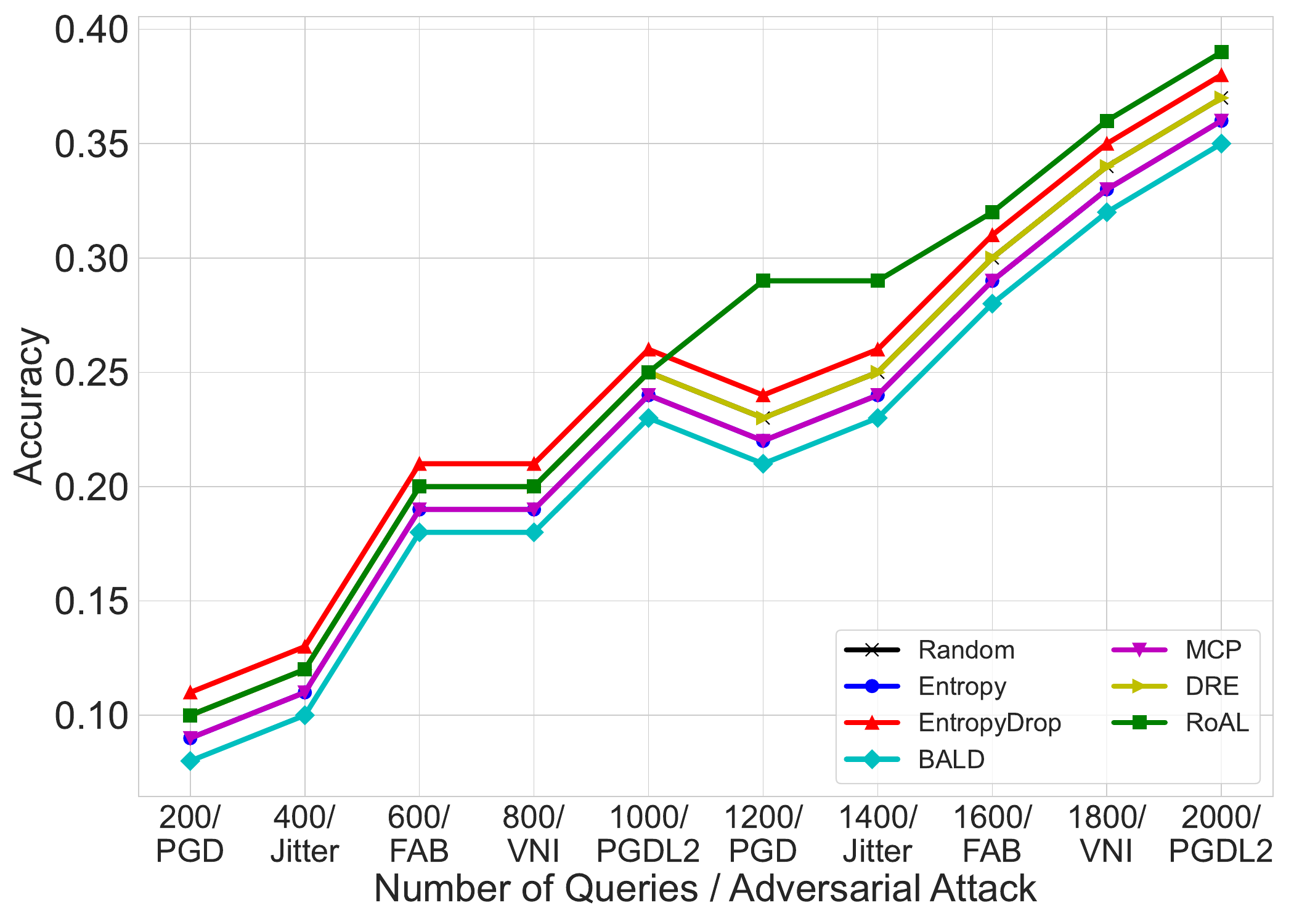}} \\
\caption{Comparative evaluation of the RoAL method and baseline methods across multiple datasets: accuracy versus the number of queries under dynamic adversarial attack.}\label{fig:Accuracy Comparison}
\end{figure*}
We evaluated the accuracy score of our RoAL method under various adversarial attacks, comparing it against several established methods across multiple datasets. Figure \ref{fig:Accuracy Comparison} provides a comprehensive visualization of these comparisons.
In the MNIST dataset, our RoAL method demonstrated superior performance, with accuracies ranging from 0.39 to 0.96, consistently outperforming other methods such as random, entropy, entropyDrop, bald, MCP, and DRE. For the Fashion dataset, the RoAL method achieved a final accuracy of 0.83, showcasing its performance against dataset complexities and adversarial manipulations. Similar trends were observed in the SVHN, CIFAR10, CIFAR100, and Caltech101 datasets, where our method matched or exceeded the performance of competing methods. Specifically, in CIFAR100 and Caltech101, our method achieved peak accuracies of 0.42 and 0.39, respectively.
\subsection{Robust Accuracy Performane}
\begin{figure*}[h!]
\centering
\subfloat[MNIST]{\includegraphics[width=.38\textwidth]{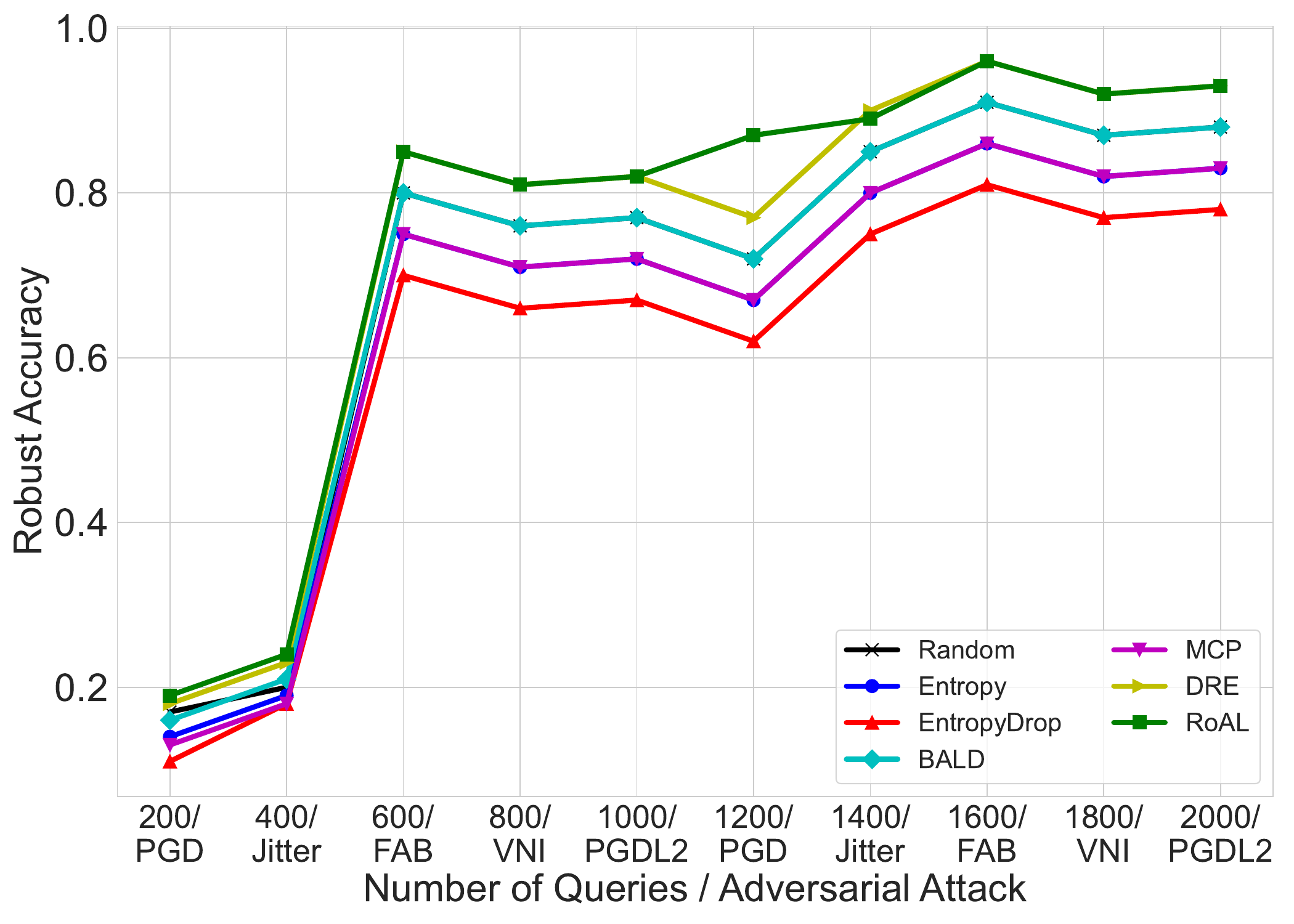}} \hspace{0.01\textwidth}
\subfloat[Fashion MNIST]{\includegraphics[width = .38\textwidth]{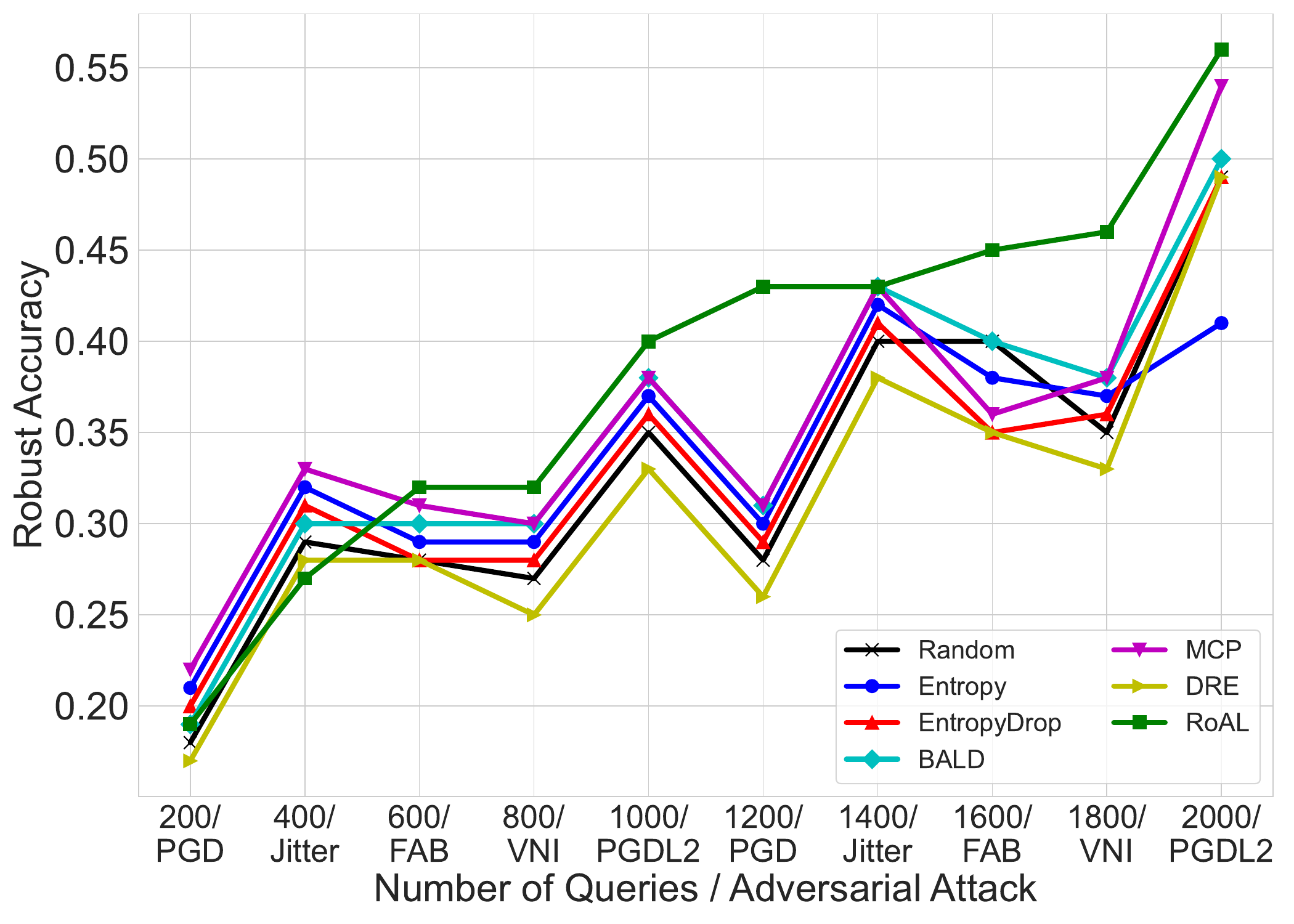}}\hspace{0.01\textwidth} \\
\subfloat[SVHN]{\includegraphics[width =.38\textwidth]{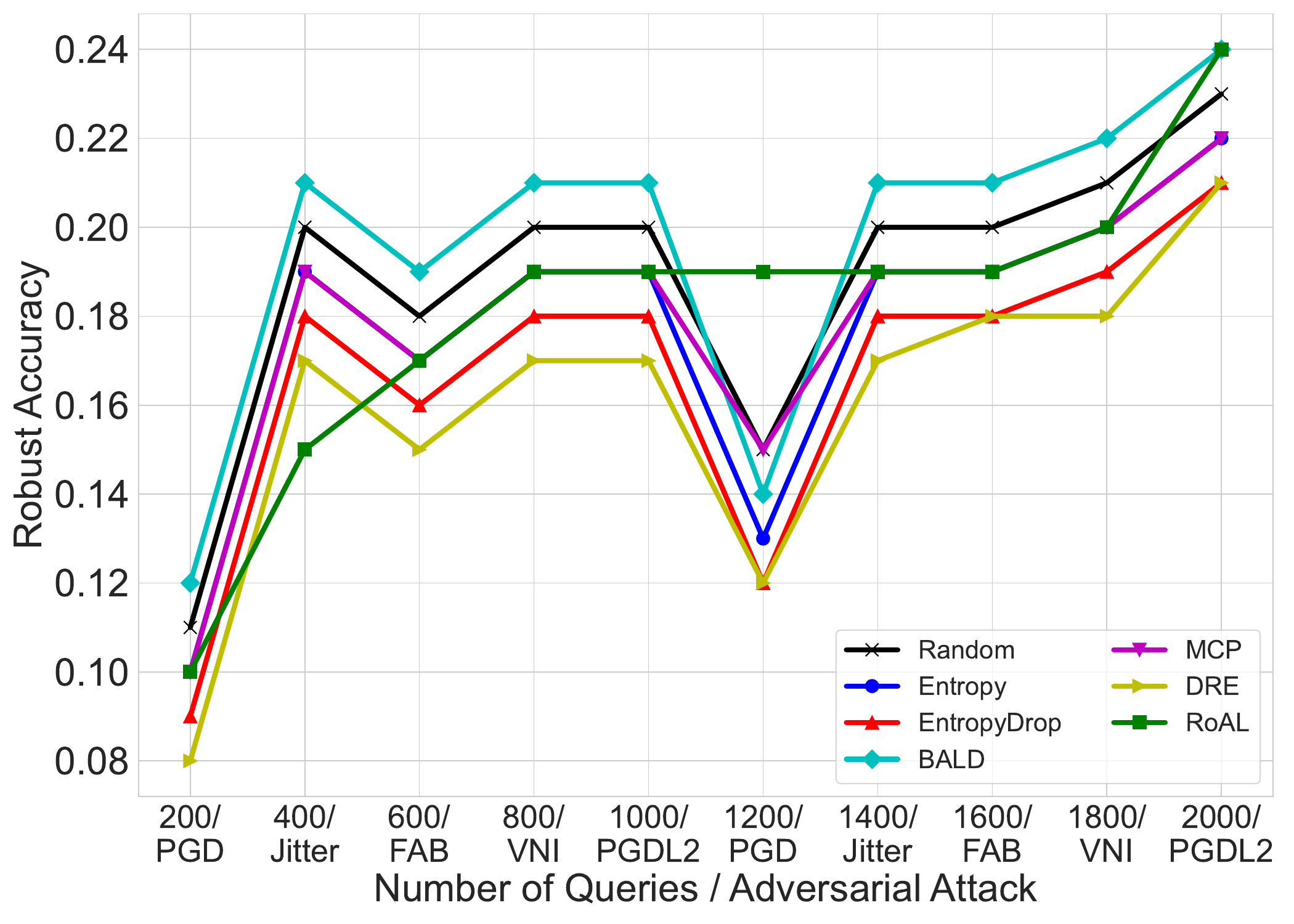}} 
\subfloat[CIFAR10]{\includegraphics[width = .38\textwidth]{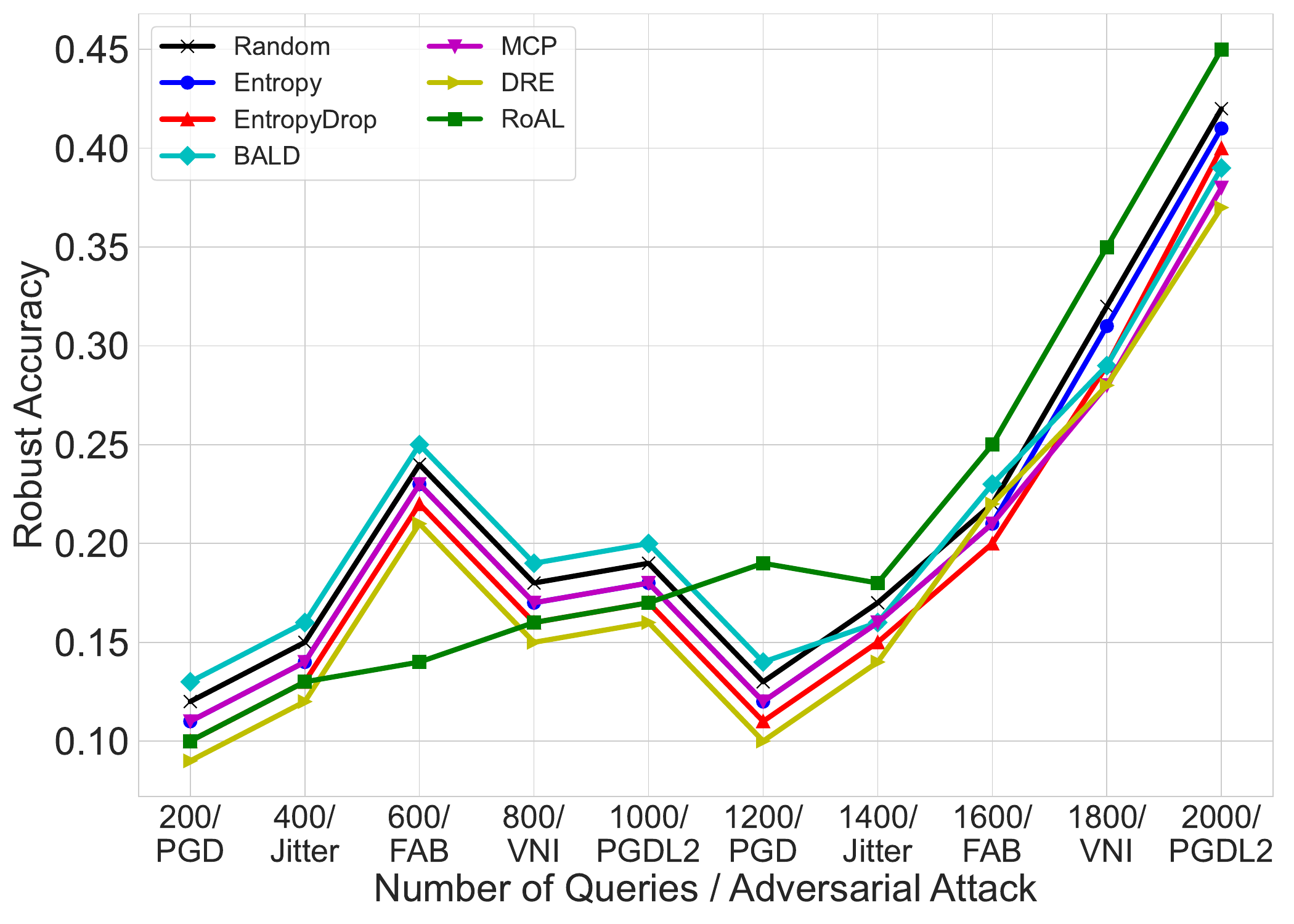}} \hspace{0.01\textwidth}\\
\subfloat[CIFAR100]{\includegraphics[width = .38\textwidth]{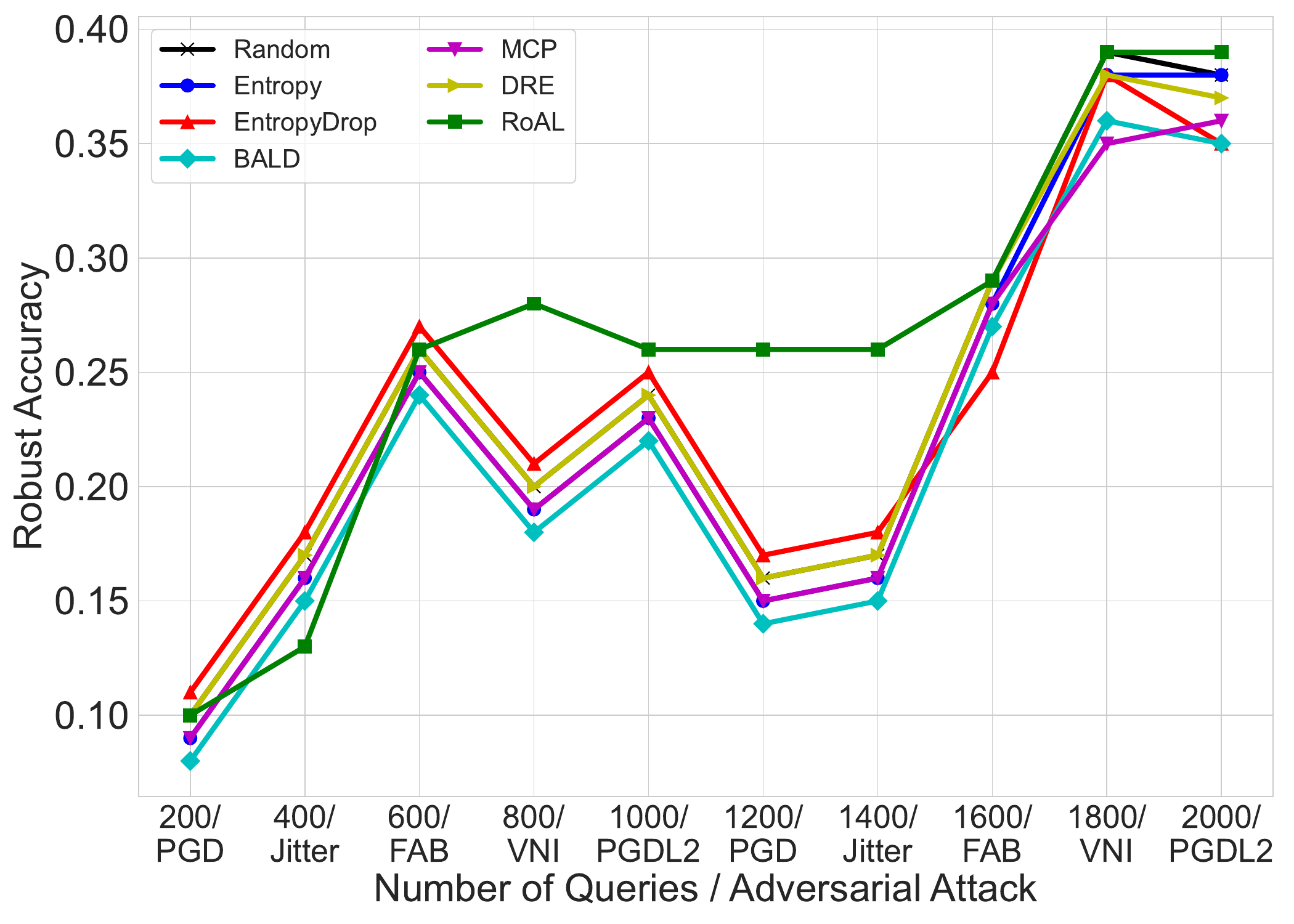}}\hspace{0.01\textwidth}
\subfloat[Caltech101]{\includegraphics[width = .38\textwidth]{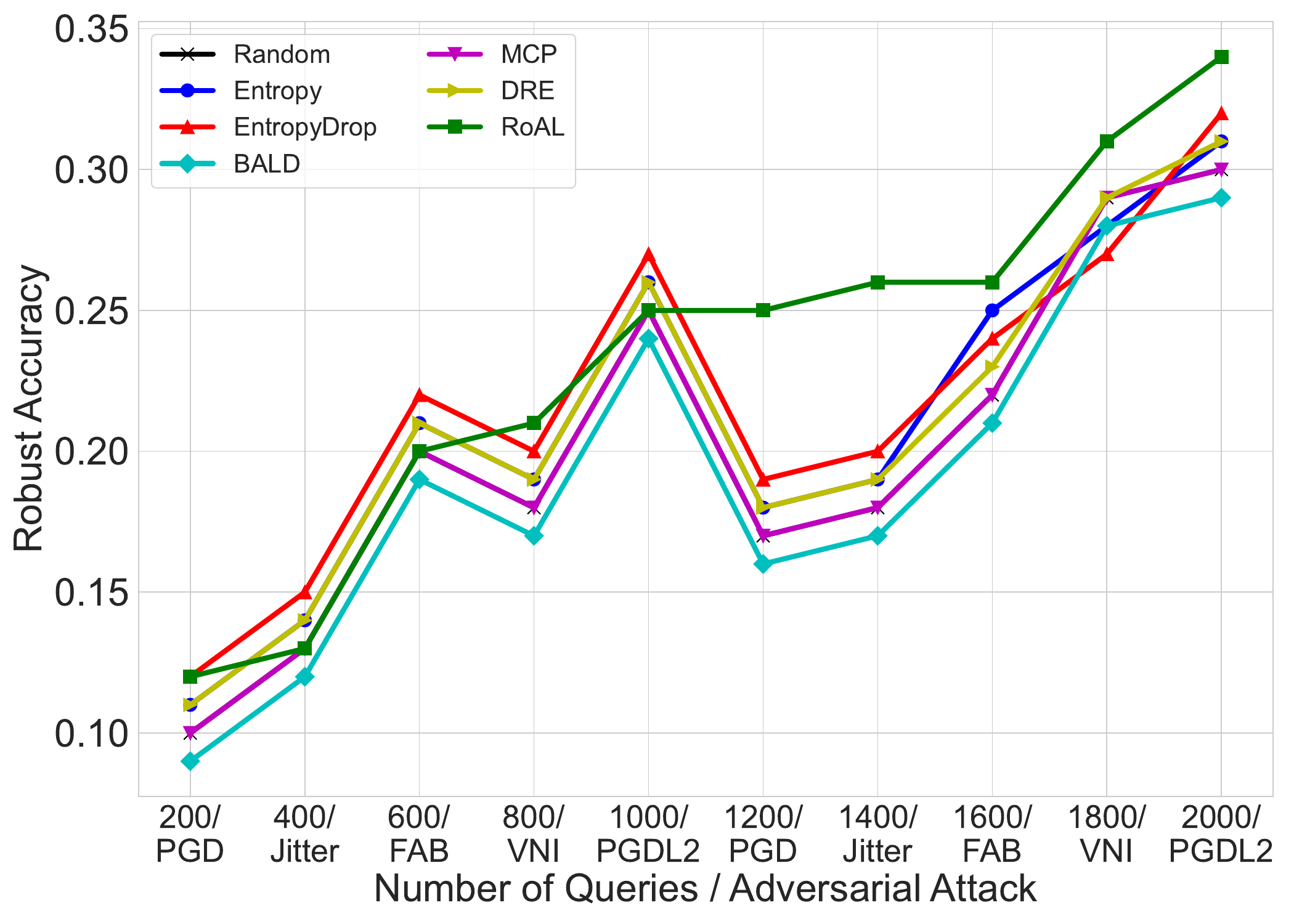}} \\
\caption{Comparative evaluation of the RoAL method and baseline methods across multiple datasets: robust accuracy versus the number of queries under dynamic adversarial attack.}\label{fig:Robust Accuracy Comparison}
\end{figure*}
The next experiment is designed to examine the robust accuracy against dynamic adversarial attacks measured by the robust accuracy score. Figure \ref{fig:Robust Accuracy Comparison} shows the robust accuracy performance of all the methods evaluated on 10 active learning iterations. The figure illustrates that across datasets, including MNIST, Fashion, and CIFAR100, the RoAL method consistently achieved the highest robust accuracy scores. The DRE method demonstrated competitive performance alongside the RoAL method in the SVHN dataset. Conversely, the MCP method emerged as the second-best performer after the RoAL method in the CIFAR10 dataset. In the Caltech101 dataset, the RoAL method again showcased superior robustness, with the DRE method being the closest competitor. 

It is interesting to see that, in the 6th iteration where all methods faced PGD adversarial attacks, all methods experienced a substantial performance drop, which could potentially be caused by catastrophic forgetting. This phenomenon occurs by the fact that the method is trained under the attack of PGDL2 on the 5th iteration while in the next iteration it changed dynamically to PGD. However, the RoAL method, augmented with EWC, exhibited stable performance under these adversarial conditions, highlighting its resilience against catastrophic forgetting. These findings emphasize the effectiveness of the RoAL method in mitigating catastrophic forgetting and enhancing robustness under adversarial conditions across diverse image classification tasks.

\subsection{Ablation Study}
In this paper, we conduct an ablation study to systematically examine the contribution of each component of the proposed model, thereby elucidating their individual and combined impacts on performance. 
\subsubsection{Robust Accuracy Under Different Acquisition Functions}
We evaluated the robust accuracy performance of various active learning acquisition functions. The experiment involved comparing uncertainty sampling in our method against four well-known acquisition functions: margin sampling, expected error reduction, cluster-based sampling, and representative sampling. Margin sampling and expected error reduction focus on selecting instances that potentially offer the most notable decrease in classification error or are near the decision boundary. In contrast, cluster-based and representative sampling method clustered instances and prioritize samples based on their uncertainty or central positioning within clusters, respectively.
\begin{table*}[htb!]
\centering
\caption{Ablation study of each acquisition function.}
\scalebox{0.7}{
\begin{tabular}{c|cccccccccc}
\hline \hline
\multirow{2}{*}{Acquisition Function} & \multicolumn{10}{c}{Robust Accuracy}                                                                                                                                                                                                                                  \\ \cline{2-11} 
                                      & \multicolumn{1}{c|}{PGD}  & \multicolumn{1}{c|}{Jitter} & \multicolumn{1}{c|}{FAB}  & \multicolumn{1}{c|}{VNI}  & \multicolumn{1}{c|}{PGDL2} & \multicolumn{1}{c|}{PGD}  & \multicolumn{1}{c|}{Jitter} & \multicolumn{1}{c|}{FAB}  & \multicolumn{1}{c|}{VNI}  & PGDL2 \\ \hline \hline
Margin Sampling                       & \multicolumn{1}{c|}{0.10} & \multicolumn{1}{c|}{0.16}   & \multicolumn{1}{c|}{0.19} & \multicolumn{1}{c|}{0.16} & \multicolumn{1}{c|}{0.24}  & \multicolumn{1}{c|}{0.18} & \multicolumn{1}{c|}{0.18}   & \multicolumn{1}{c|}{0.24} & \multicolumn{1}{c|}{0.35} & 0.48  \\ \hline
Expected Error Reduction              & \multicolumn{1}{c|}{0.28} & \multicolumn{1}{c|}{0.14}   & \multicolumn{1}{c|}{0.16} & \multicolumn{1}{c|}{0.19} & \multicolumn{1}{c|}{0.27}  & \multicolumn{1}{c|}{0.29} & \multicolumn{1}{c|}{0.24}   & \multicolumn{1}{c|}{0.26} & \multicolumn{1}{c|}{0.28} & 0.29  \\ \hline
Cluster Based Sampling                & \multicolumn{1}{c|}{0.10} & \multicolumn{1}{c|}{0.10}   & \multicolumn{1}{c|}{0.12} & \multicolumn{1}{c|}{0.12} & \multicolumn{1}{c|}{0.16}  & \multicolumn{1}{c|}{0.20} & \multicolumn{1}{c|}{0.24}   & \multicolumn{1}{c|}{0.24} & \multicolumn{1}{c|}{0.26} & 0.35  \\ \hline
Representative Sampling               & \multicolumn{1}{c|}{0.17} & \multicolumn{1}{c|}{0.17}   & \multicolumn{1}{c|}{0.17} & \multicolumn{1}{c|}{0.18} & \multicolumn{1}{c|}{0.10}  & \multicolumn{1}{c|}{0.18} & \multicolumn{1}{c|}{0.21}   & \multicolumn{1}{c|}{0.20} & \multicolumn{1}{c|}{0.20} & 0.33  \\ \hline
Uncertainty Sampling *                & \multicolumn{1}{c|}{0.19} & \multicolumn{1}{c|}{0.27}   & \multicolumn{1}{c|}{0.32} & \multicolumn{1}{c|}{0.32} & \multicolumn{1}{c|}{0.4}   & \multicolumn{1}{c|}{0.43} & \multicolumn{1}{c|}{0.43}   & \multicolumn{1}{c|}{0.45} & \multicolumn{1}{c|}{0.41} & 0.56  \\ \hline \hline
\end{tabular}}\label{tab:ablation}
\end{table*}
Our method, employing uncertainty sampling, consistently selected instances where the classifier showed the highest uncertainty in classification. This approach often included instances with low confidence scores or those near decision boundaries, which proved effective in maintaining robust accuracy under various adversarial conditions. Notably, uncertainty sampling achieved robust accuracy scores as high as 0.56 under PGDL2 attack conditions, substantially outperforming other methods which showed lower scores such as 0.48, 0.29, 0.35, and 0.33 for margin sampling, expected error reduction, cluster-based sampling, and representative sampling respectively under similar conditions. These results highlight the effectiveness of uncertainty sampling in enhancing the robustness of the active learning process, as summarized in Table \ref{tab:ablation}.
\subsubsection{Robust Accuracy Under Different Initial Labeled Data}
\begin{figure*}[htb!]
\centering
\subfloat[MNIST]{\includegraphics[width=.38\textwidth]{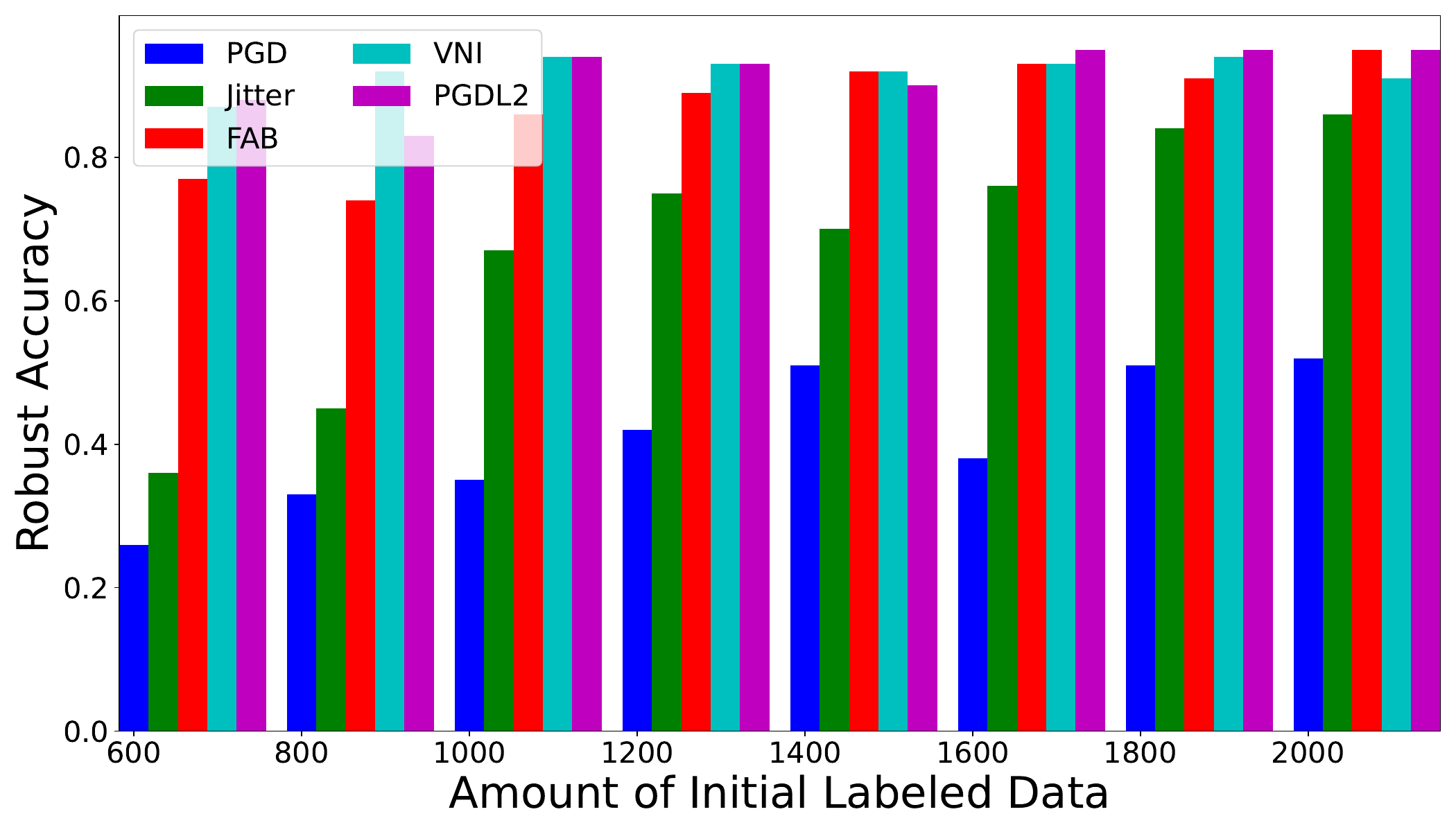}} \hspace{0.01\textwidth}
\subfloat[Fashion MNIST]{\includegraphics[width = .38\textwidth]{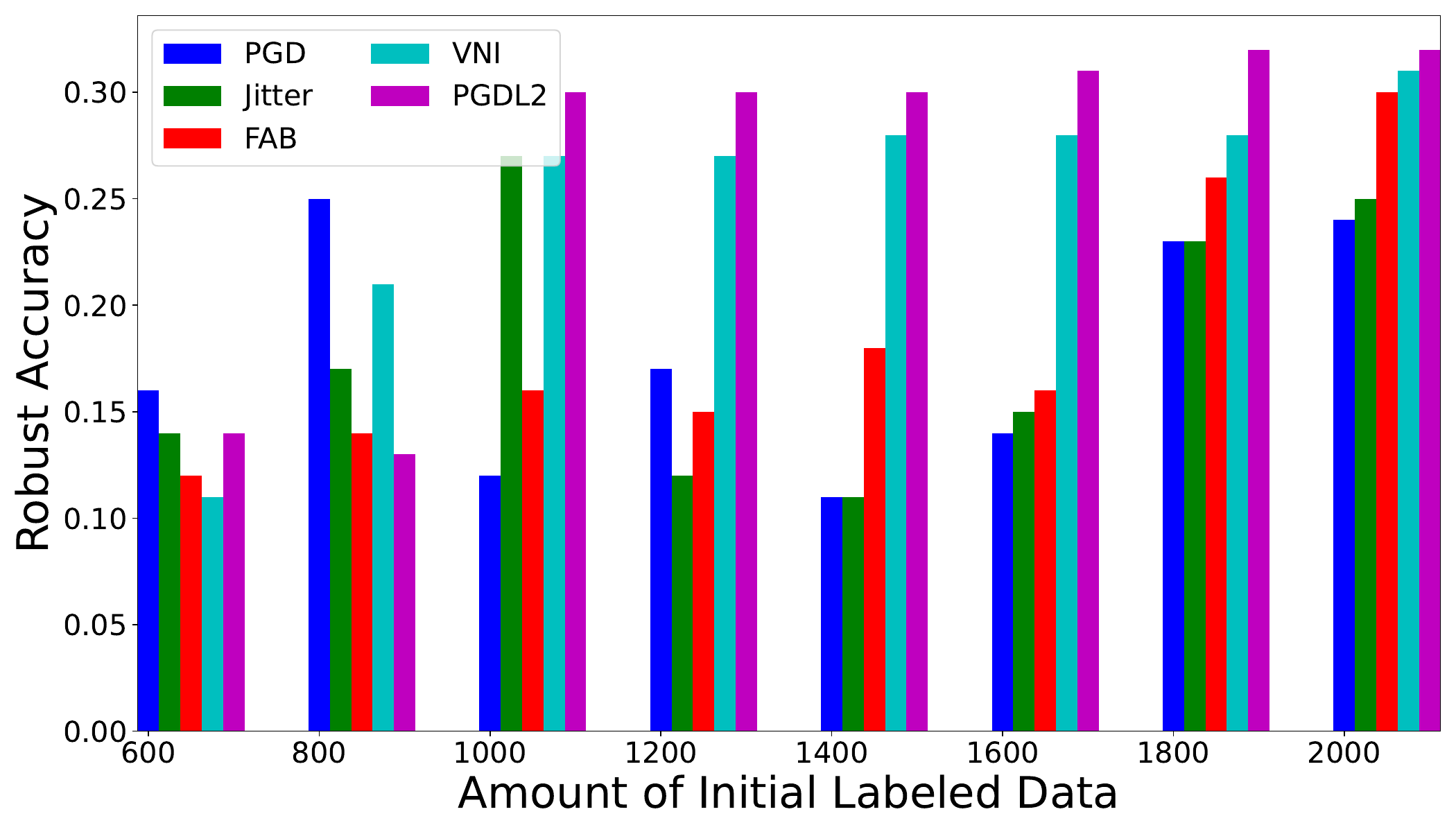}}\hspace{0.01\textwidth} \\
\subfloat[CALTECH101]{\includegraphics[width =.38\textwidth]{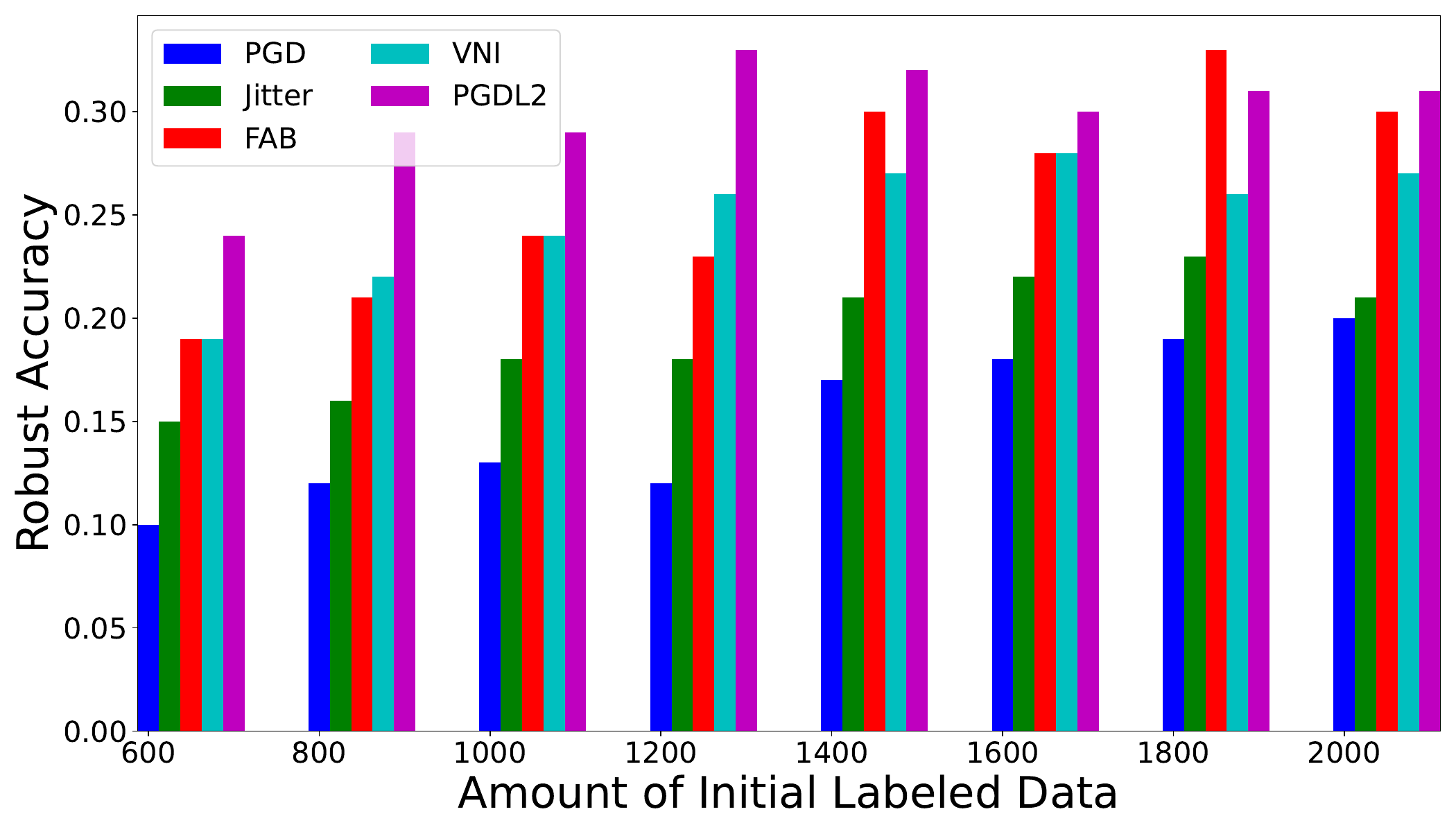}} 
\subfloat[CIFAR10]{\includegraphics[width = .38\textwidth]{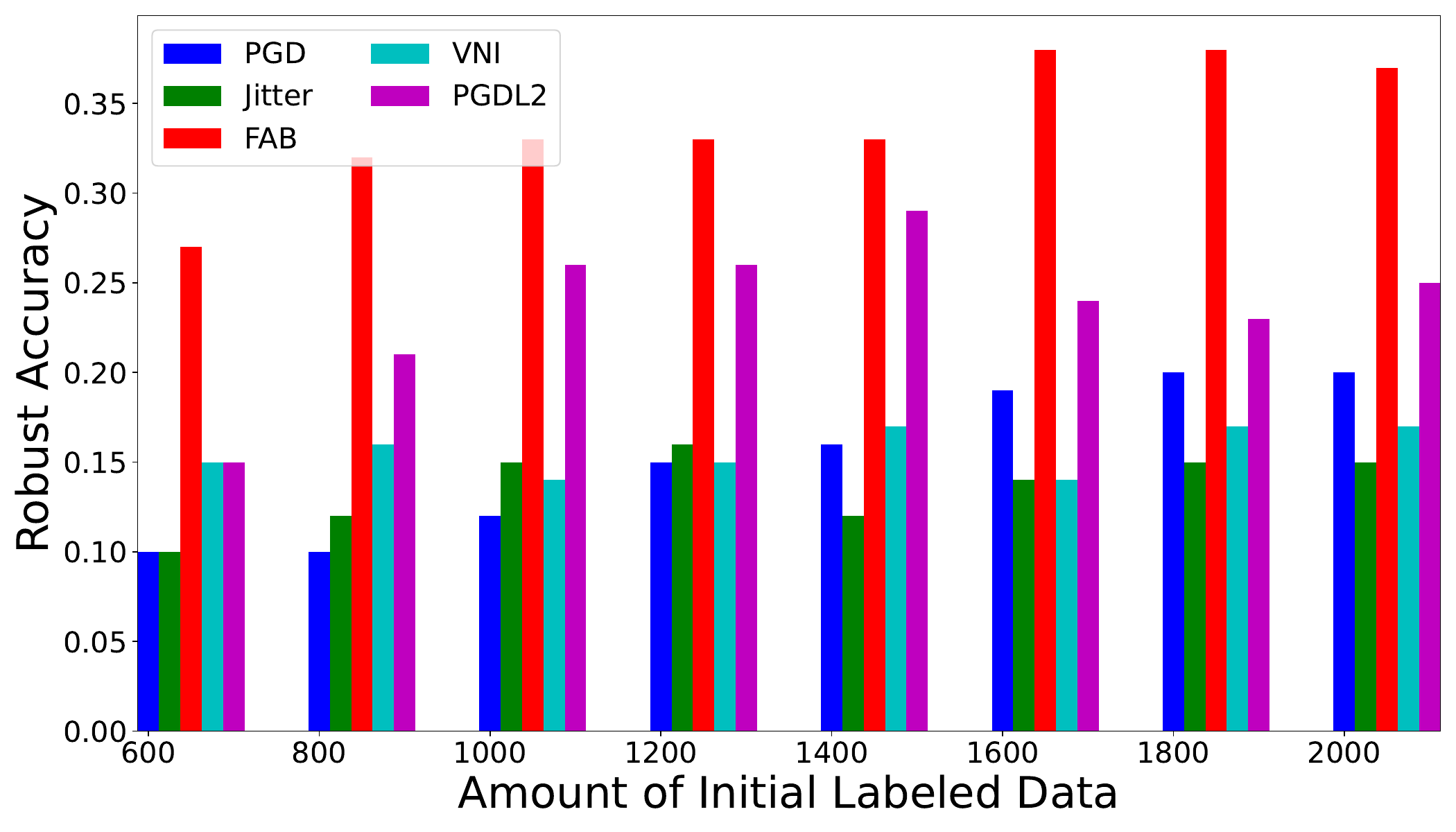}} \hspace{0.01\textwidth}\\
\caption{Comparative evaluation of the RoAL under different amount of initial labeled data.}\label{fig:experiment initial data}
\end{figure*}
In the next study, we investigate the performance of the RoAL approach as it relates to varying quantities of initial labeled data. The amount of initial labeled data is crucial for effectively training the initial model in an active learning framework. Figure \ref{fig:experiment initial data} demonstrates the impact of initial data volume on the robustness of accuracy against adversarial attacks, with datasets varying and data amounts ranging from 600 to 2000. This experiment reveals a consistent trend of gradual improvement in robust accuracy across various datasets, exemplified by the MNIST dataset where an increase in initial labeled data correlates with enhanced robustness. Particular noteworthy is the performance of the RoAL method under the PGDL2 attack, which consistently exhibits higher robustness compared to other adversarial attacks such as PGD, Jitter, FAB, and VNI. This pattern is not only observable in the MNIST dataset but also mirrors similar trends in Fashion MNIST, CIFAR-10, and Caltech101 datasets. In these studies, the RoAL method under the PGDL2 attack demonstrated superior robustness, followed closely by VNI, with other adversarial attacks trailing. Overall, the findings show that an increase in the initial labeled data significantly bolsters the robustness of the RoAL method against adversarial attacks. 
\subsubsection{The effect of $\lambda$ in EWC regularization}
The RoAL method has one hyperparameter i.e. $\lambda$ to control the EWC regularization value. The EWC simply works by adding a regularizer ($\lambda$) to the loss function that prevents the most important weights from deviating far from the consolidated values during the learning tasks. It is widely known \citep{Kutalev2021StabilizingEW} that if $\lambda$ is small, the resistance to change weights, in general, will be small, and during sequential learning, the neural network will learn the current task better, but it will forget the knowledge from previous learned tasks faster. Conversely, if $\lambda$ is too large, then the resistance to change weights will also be large, and the network will retain the previous knowledge learned on past tasks well, but will be slow in learning a new task. 
\begin{table*}[h]
\centering
\caption{The effect of $\lambda$ in EWC regularization.}
\scalebox{0.9}{
\begin{tabular}{c|cccccccccc}
\hline \hline
\multirow{2}{*}{$\lambda$ Value} & \multicolumn{10}{c}{Robust Accuracy}                                                                                                                                                                                                                                  \\ \cline{2-11} 
                              & \multicolumn{1}{c|}{PGD}  & \multicolumn{1}{c|}{Jitter} & \multicolumn{1}{c|}{FAB}  & \multicolumn{1}{c|}{VNI}  & \multicolumn{1}{c|}{PGDL2} & \multicolumn{1}{c|}{PGD}  & \multicolumn{1}{c|}{Jitter} & \multicolumn{1}{c|}{FAB}  & \multicolumn{1}{c|}{VNI}  & PGDL2 \\ \hline \hline
0.1                           & \multicolumn{1}{c|}{0.12} & \multicolumn{1}{c|}{0.16}   & \multicolumn{1}{c|}{0.20}  & \multicolumn{1}{c|}{0.20}  & \multicolumn{1}{c|}{0.20}   & \multicolumn{1}{c|}{0.16} & \multicolumn{1}{c|}{0.15}   & \multicolumn{1}{c|}{0.20}  & \multicolumn{1}{c|}{0.23} & 0.23  \\ \hline
0.2                           & \multicolumn{1}{c|}{0.10} & \multicolumn{1}{c|}{0.13}   & \multicolumn{1}{c|}{0.25} & \multicolumn{1}{c|}{0.21} & \multicolumn{1}{c|}{0.19}  & \multicolumn{1}{c|}{0.20} & \multicolumn{1}{c|}{0.16}   & \multicolumn{1}{c|}{0.20} & \multicolumn{1}{c|}{0.23} & 0.23  \\ \hline
0.3                           & \multicolumn{1}{c|}{0.10} & \multicolumn{1}{c|}{0.1}    & \multicolumn{1}{c|}{0.25} & \multicolumn{1}{c|}{0.21} & \multicolumn{1}{c|}{0.21}  & \multicolumn{1}{c|}{0.16} & \multicolumn{1}{c|}{0.14}   & \multicolumn{1}{c|}{0.24} & \multicolumn{1}{c|}{0.19} & 0.24  \\ \hline
0.4                           & \multicolumn{1}{c|}{0.10} & \multicolumn{1}{c|}{0.14}   & \multicolumn{1}{c|}{0.21} & \multicolumn{1}{c|}{0.21} & \multicolumn{1}{c|}{0.21}  & \multicolumn{1}{c|}{0.17} & \multicolumn{1}{c|}{0.14}   & \multicolumn{1}{c|}{0.21} & \multicolumn{1}{c|}{0.26} & 0.24  \\ \hline
0.5                           & \multicolumn{1}{c|}{0.10} & \multicolumn{1}{c|}{0.13}   & \multicolumn{1}{c|}{0.22} & \multicolumn{1}{c|}{0.16} & \multicolumn{1}{c|}{0.17}  & \multicolumn{1}{c|}{0.19} & \multicolumn{1}{c|}{0.20}   & \multicolumn{1}{c|}{0.25} & \multicolumn{1}{c|}{0.35} & 0.45  \\ \hline
0.6                           & \multicolumn{1}{c|}{0.10} & \multicolumn{1}{c|}{0.10}   & \multicolumn{1}{c|}{0.24} & \multicolumn{1}{c|}{0.20} & \multicolumn{1}{c|}{0.19}  & \multicolumn{1}{c|}{0.19} & \multicolumn{1}{c|}{0.21}   & \multicolumn{1}{c|}{0.24} & \multicolumn{1}{c|}{0.34} & 0.46  \\ \hline
0.7                           & \multicolumn{1}{c|}{0.10} & \multicolumn{1}{c|}{0.11}   & \multicolumn{1}{c|}{0.30} & \multicolumn{1}{c|}{0.24} & \multicolumn{1}{c|}{0.23}  & \multicolumn{1}{c|}{0.23} & \multicolumn{1}{c|}{0.20}   & \multicolumn{1}{c|}{0.24} & \multicolumn{1}{c|}{0.34} & 0.46  \\ \hline
0.8                           & \multicolumn{1}{c|}{0.10} & \multicolumn{1}{c|}{0.10}   & \multicolumn{1}{c|}{0.28} & \multicolumn{1}{c|}{0.24} & \multicolumn{1}{c|}{0.21}  & \multicolumn{1}{c|}{0.22} & \multicolumn{1}{c|}{0.20}   & \multicolumn{1}{c|}{0.26} & \multicolumn{1}{c|}{0.36} & 0.47  \\ \hline
0.9                           & \multicolumn{1}{c|}{0.10} & \multicolumn{1}{c|}{0.10}   & \multicolumn{1}{c|}{0.20} & \multicolumn{1}{c|}{0.20} & \multicolumn{1}{c|}{0.19}  & \multicolumn{1}{c|}{0.19} & \multicolumn{1}{c|}{0.20}   & \multicolumn{1}{c|}{0.24} & \multicolumn{1}{c|}{0.34} & 0.48  \\ \hline
1.0                           & \multicolumn{1}{c|}{0.10} & \multicolumn{1}{c|}{0.09}   & \multicolumn{1}{c|}{0.24} & \multicolumn{1}{c|}{0.20} & \multicolumn{1}{c|}{0.16}  & \multicolumn{1}{c|}{0.19} & \multicolumn{1}{c|}{0.20}   & \multicolumn{1}{c|}{0.21} & \multicolumn{1}{c|}{0.37} & 0.47  \\ \hline \hline
\end{tabular}}\label{tab:ablation_lambda}
\end{table*}
Therefore, we conduct the experimental study to select the best $\lambda$ value. For this experiment, we use CIFAR10 dataset and varying the $\lambda$ score for each experiment. Table \ref{tab:ablation_lambda} illustrates the robust accuracy performance of varying $\lambda$ values. At lower $\lambda$ values, such as 0.1 to 0.3, the network's performance is somewhat inconsistent, which suggests it’s adapting quickly to new tasks but forgetting previous learning just as fast. On the other hand, increasing $\lambda$ to higher values, like above 0.5, results in better robust accuracy, particularly against attacks like VNI and PGDL2, where robust accuracy peaks at 0.47 for $\lambda$=1.0. This improvement indicates stronger retention of past knowledge, though it might slow down the network in learning new tasks. Mid-range $\lambda$ values, around 0.5 to 0.7, show a better balance, showing good robust accuracy that points to an effective compromise between retaining previous information and learning new tasks. 

\section{Complexity Analysis}
The computational complexity of the Robust Active Learning (RoAL) method with Elastic Weight Consolidation (EWC) can be approximated as $O(T \cdot (b \cdot m + p))$, where $T$ represents the number of training iterations, $b$ denotes the number of batches processed per iteration, and $m$ reflects the complexity of processing each batch, including tasks such as loss computation, EWC application, and parameter updates. The term $p$ encompasses the computational load of additional crucial operations like updating model parameters ($\theta^*$), recalculating Fisher information, and evaluating the model under both normal and adversarial conditions. While the algorithm demonstrates linear scalability with respect to the number of batches and a direct correlation to the volume of data and batch size, it is important to recognize potential bottlenecks in $p$, which could impact the overall computational complexity or performance of the algorithm.
\section{Discussion}
The results show the effectiveness of the proposed RoAL method in achieving robust accuracy for active learning under dynamic adversarial threats. To better understand the significance of these findings, we discuss them in further detail in this section as compared to other baseline methods in the active learning domain. 

Firstly, as we discussed the proposed method performs better in terms of robust accuracy. We, therefore,highlight the improvement of the proposed RoAL method in terms of robust accuracy as compared to the best two state-of-the-art methods, i.e., DRE \citep{DRE} and MCP \citep{MCP}. Table \ref{tab:percentage_improvement} shows the percentage of improvement of RoAL as compared to DRE and MCP in terms of robust accuracy score. 
Overall, on average, RoAL outperformed DRE by 10.88\% and MCP by 10.82\% based on the considered datasets.
\begin{table}[ht!]
\centering
\caption{ Percentage Improvement of RoAL with respect to two best baselines}
\label{tab:percentage_improvement}
\begin{tabular}{c|cc}
\hline \hline
\multirow{2}{*}{Dataset} & \multicolumn{2}{c}{\% of Improvement}         \\ \cline{2-3} 
                         & \multicolumn{1}{c|}{RoAL vs DRE} & RoAL vs MCP \\ \hline \hline
MNIST                    & \multicolumn{1}{c|}{0}           & 12.05       \\ \hline
Fashion MNIST            & \multicolumn{1}{c|}{14.29}       & 3.70        \\ \hline
Cifar10                  & \multicolumn{1}{c|}{21.62}       & 18.42       \\ \hline
Cifar100                 & \multicolumn{1}{c|}{5.41}        & 8.33        \\ \hline
SVHN                     & \multicolumn{1}{c|}{14.29}       & 9.09        \\ \hline
Caltech101               & \multicolumn{1}{c|}{9.68}        & 13.33       \\ \hline 
\textbf{Average Improvement}      & \multicolumn{1}{c|}{\textbf{10.88}}          & \textbf{10.82}              \\ \hline \hline
\end{tabular}
\end{table}
Secondly, DRE and MCP methods are recent additions to the active learning field, specifically designed to counter adversarial attacks. While EADA combats these attacks by training deep neural networks with perturbed data, MCP improves model retraining by clustering test samples into boundary areas based on predicted classes. Although both approaches enhance active learning accuracy, they often do so at the expense of robust accuracy due to catastrophic forgetting caused by dynamic adversarial attacks. In contrast, RoAL maintains robust accuracy by implementing EWC, thereby reducing catastrophic forgetting and boosting robust accuracy. This improvement highlights the advantages of the proposed RoAL method over recent state-of-the-art methods like DRE and MCP.
\section{Conclusion and Future Work}\label{section:conclusion}
In conclusion, the contributions of this paper substantially enhance the field of active learning under adversarial attack. We introduced a dynamic adversarial attack model tailored specifically for active learning environments, which more closely mimics real-world adversarial behaviors. This model has improved our understanding of the practical challenges that active learning systems may face, equipping us with the knowledge to develop more resilient approaches. Additionally, we proposed a novel method that integrates EWC with active learning, specifically designed to solve catastrophic forgetting in scenarios resulting from dynamic adversarial threats. Our approach not only mitigates the impact of these threats but also preserves learning performance over successive iterations. The efficacy of our method was confirmed through extensive experimental evaluations, which demonstrated that optimizing active learning with EWC and uncertainty sampling notably bolsters the model’s robustness against dynamic adversarial attacks. This ensures stable and reliable performance across a variety of learning tasks. Furthermore, our ablation study indicated that a $\lambda$ value of 0.5, coupled with uncertainty sampling, critically supports the performance enhancements of our RoAL method. The findings from this study could significantly influence the development of more robust active learning frameworks that can effectively handle adversarial conditions in real-world applications.

Finally, the successful application of EWC in this study opens the door for investigating a broader range of continual learning approaches beyond EWC. Future research could delve into different continual learning mechanisms, such as synaptic intelligence, gradient episodic memory, or experience replay, to evaluate their effectiveness in active learning environments facing adversarial attacks. This exploration would involve testing these methods under similar conditions to compare their ability to prevent catastrophic forgetting and enhance model robustness. Additionally, experiments could be designed to combine these approaches with various active learning strategies to assess potential synergies. Such studies will not only provide a comprehensive understanding of the comparative strengths and weaknesses of each continual learning method but also contribute to refining and innovating more resilient active learning frameworks capable of overcoming sophisticated adversarial challenges.
\bibliographystyle{elsarticle-num} 
\bibliography{references}


\section*{Credit authorship contribution statement}

\textbf{Ricky M. Fajri:} Conceptualization, Methodology, Software, Investigation, Writing - Original Draft. \textbf{Yulong Pei:} Conceptualization, Methodology, Writing - Review \& Editing. \textbf{Lu Yin:} Conceptualization, Methodology, Writing - Review \& Editing. \textbf{Mykola Pechenizkiy:} Methodology, Supervision.

\section*{Declaration of Competing Interest}

The authors declare that they have no known competing financial interests or personal relationships that could have appeared to influence the work reported in this paper.

\section*{Funding}
The authors did not receive support from any organization for this work.
\end{document}